\documentclass[10pt,twocolumn,letterpaper]{article}

\usepackage[pagenumbers]{cvpr} 

\definecolor{cvprblue}{rgb}{0.21,0.49,0.74}
\usepackage[pagebackref,breaklinks,colorlinks,allcolors=cvprblue]{hyperref}

\usepackage[utf8]{inputenc} 
\usepackage[T1]{fontenc}    
\usepackage{url}            
\usepackage{booktabs}       
\usepackage{amsfonts}       
\usepackage{nicefrac}       
\usepackage{microtype}      
\usepackage{xcolor}         
\usepackage{amsmath}

\usepackage{bm}
\usepackage{setspace}
\usepackage{hyperref}       
\usepackage{wrapfig}
\usepackage{graphicx}	
\usepackage{amsmath}
\usepackage{algorithm}
\usepackage{algorithmic}
\usepackage{amssymb}	
\usepackage{booktabs}
\usepackage{times}
\usepackage{epsfig}
\usepackage{multirow, bigstrut}
\usepackage{caption}
\usepackage{float}
\usepackage{placeins}
\usepackage{amsmath}
\usepackage{color, colortbl}
\usepackage{stfloats}
\usepackage{enumitem}
\usepackage{tabularx}
\usepackage{xstring}
\usepackage{xspace}
\usepackage{url}
\usepackage{subcaption}
\usepackage{overpic}
\usepackage[hang,flushmargin]{footmisc}
\definecolor{color3}{gray}{0.95}
\definecolor{lightblue}{HTML}{1E90FF} 

\definecolor{lightlightblue}{HTML}{B0E0E6} 
\definecolor{lightred}{HTML}{B22222} 
\definecolor{gree}{HTML}{A8E0B7} 
\definecolor{rouse}{rgb}{0.981,0.961,0.941}
\definecolor{lightgreen}{rgb}{0.9, 0.99, 0.9}
\definecolor{light-yellow}{rgb}{1,1,0.93}
\definecolor{lightgray5}{gray}{0.98}
\definecolor{lightgray6}{gray}{0.96}

\newcommand{\h}{0}
\newcommand{\w}{0.15}
\newcommand{\wa}{0.15}
\newlength \g
\newcommand{\name}{0}
\usepackage{adjustbox}

\let\OLDthebibliography\thebibliography
\renewcommand\thebibliography[1]{
  \OLDthebibliography{#1}
  \setlength{\parskip}{0pt}
  \setlength{\itemsep}{0pt plus 0.3ex}
}

\def\paperID{17004} 
\def\confName{CVPR}
\def\confYear{2025}

\title{Cross-Scan Mamba with Masked Training for Robust Spectral Imaging}

\author{%
	Wenzhe Tian $^{1,*}$, Haijin Zeng $^{2,}$\thanks{Equal Contribution, $\dagger$ Corresponding Author} ~, Yin-Ping Zhao $^3$, Yongyong Chen $^{4,\dagger}$, \\ Zhen Wang $^3$, Xuelong Li $^3$ \\
	$^{1}$ University of Helsinki, $^2$  IMEC \& Universiteit Gent, \\ $^{3}$ Northwestern Polytechnical University, $^4$ 	Harbin Institute of Technology (Shenzhen)
}

\begin{document}
\maketitle
\begin{abstract}
Snapshot Compressive Imaging (SCI) enables fast spectral imaging but requires effective decoding algorithms for hyperspectral image (HSI) reconstruction from compressed measurements. Current CNN-based methods are limited in modeling long-range dependencies, while Transformer-based models face high computational complexity. Although recent Mamba models outperform CNNs and Transformers in RGB tasks concerning computational efficiency or accuracy, they are not specifically optimized to fully leverage the local spatial and spectral correlations inherent in HSIs.
To address this, we propose the \textbf{C}ross-\textbf{S}canning Mamba, named CS-Mamba, that employs a Spatial-Spectral SSM for global-local balanced context encoding and cross-channel interaction promotion.
Besides, while current reconstruction algorithms perform increasingly well in simulation scenarios, they exhibit suboptimal performance on real data due to limited generalization capability. During the training process, the model may not capture the inherent features of the images but rather learn the parameters to mitigate specific noise and loss, which may lead to a decline in reconstruction quality when faced with real scenes. To overcome this challenge, we propose a masked training method to enhance the generalization ability of models.
Experiment results show that our CS-Mamba achieves state-of-the-art performance and the masked training method can better reconstruct smooth features to improve the visual quality.
\end{abstract}

\section{Introduction}
\label{sec:intro}

\begin{figure}[t]
  \centering
   \includegraphics[width=0.98\linewidth]{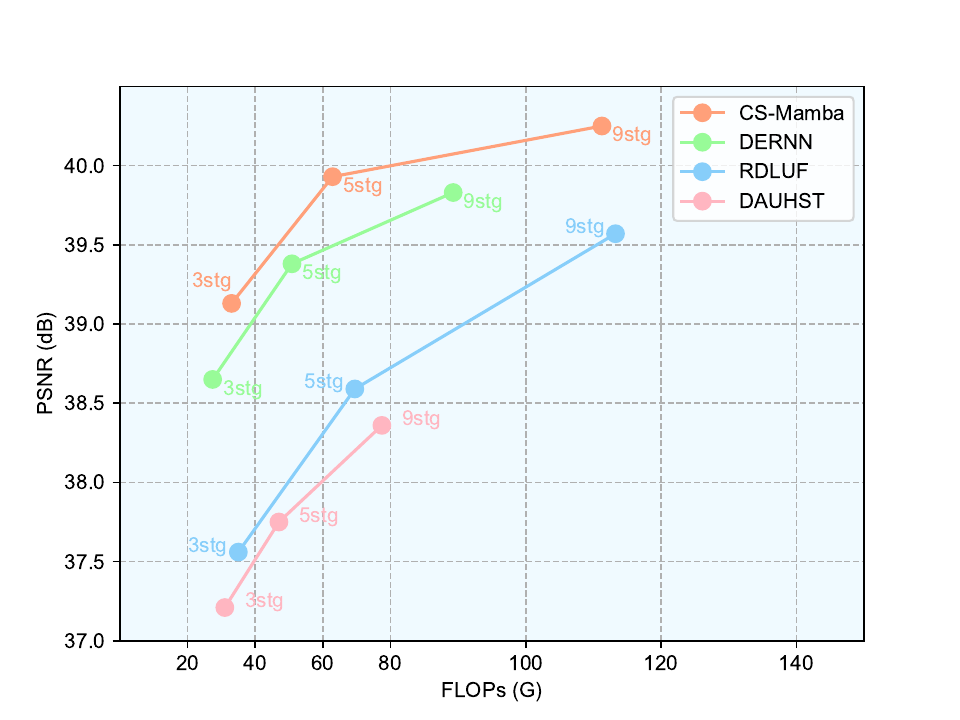}

   \vspace{-4mm}
   \caption{PSNR-FLOPs comparisons of CS-Mamba and previous Deep Unfolding SOTA methods.}
    \vspace{-5mm}
   \label{fig:teaser}
\end{figure}

Hyperspectral images (HSIs) contain a large number of spectral bands to store rich detail information and thus are widely used in various fields such as object tracking \cite{ot_1,ot_2,ot_3}, remote sensing \cite{chen2017denoising, he2020non, yuan2018hyperspectral},  super-resolution \cite{dong2021model, sr_2, sr_3}, and medical image processing \cite{mi_1,mi_2,mi_3}.
Traditional imaging systems used for collecting HSIs involve scanning along the spatial or spectral dimensions but this imaging process is time-consuming and can not effectively capture dynamic objects. 
Recently, snapshot compressive imaging (SCI) systems have been developed \cite{sci_1, sci_2} as an alternative method for obtaining HSIs at video rate. 
SCI systems capture multiple image bands in a single snapshot, compressing the information of HSIs into a 2D measurement. \cite{sci_3, sci_4, sci_5, sci_6}. 
Among these systems, Coded Aperture Snapshot Spectral Imaging (CASSI) has emerged and become mainstream~\cite{gap_tv,icme_sci_1}. 
CASSI utilizes a physical mask to modulate signals from HSIs, shifting these modulated images to different spatial locations on the detector plane via a disperser~\cite{mst,icme_sci_2}. 
However, the performance of the CASSI system heavily depends on the efficiency of the reconstruction algorithm.

In recent years, numerous reconstruction algorithms have been proposed, including model-based methods and deep learning-based methods.
Traditional model-based methods typically attempt to model the inverse process of image reconstruction, using manually designed priors to derive the solution to an optimization problem.
Although these methods are highly interpretable, they often require manual parameter tuning and tend to be slow. 
With the success of deep learning methods in various visual tasks, CNN and Transformer-based methods have been developed and applied to HSI reconstruction, achieving higher image reconstruction quality and speed than traditional algorithms. 
The CNN-based methods~\cite{lambda,tsa_net} have an advantage in extracting local features from images, but they lack the ability to model long-range dependencies effectively. 
Transformer-based methods~\cite{mst,mst_pp,dauhst,cst} realize a global receptive field, resulting in superior performance compared to CNNs but higher computational costs. 
Although some mechanisms can mitigate this issue, they do not fundamentally change the quadratic complexity problem caused by the attention mechanism in the Transformer architecture.

The limitation of Transformers in spectral SCI remains unresolved. However, in the realm of large language model and RGB image vision, an advanced solution for this problem exists, known as Mamba~\cite{gu2023mamba}.
Through data-dependent selective SSMs with parallel scanning and hardware-wise strategies, Mamba establishes a global receptive field and achieves near-linear complexity.
Mamba-based methods have been shown to outperform previous methods in various tasks due to their excellent long sequence modeling capabilities and lower computational costs~\cite{mamba_survey,vmamba2,mambair,lightm-unet}. 

Subsequently, many works have begun to explore the application and performance of the Mamba model in various visual tasks. Representative works, such as the visual state space model~\cite{shi2024vmambair} and vision Mamba~\cite{vision_mamba}, have demonstrated Mamba's superiority for image processing, particularly in obtaining fewer parameters and faster speed. However, most Mamba-based models only conduct the scanning process along the spatial dimension, lacking cross-channel interaction, despite the importance of spectral information for HSI reconstruction. 
Recently, some Mamba-based methods have made progress in developing an SSM mechanism along the channel (spectral) dimension for certain visual tasks~\cite{s2mamba,huang2024spectral}, achieving better results compared to scanning only along the spatial dimension. \emph{However, this scanning method does not consider the inherently local features of HSIs in both spatial and spectral dimensions.}
Additionally, previous reconstruction algorithms have achieved better results in experiments by continuously improving the backbone and framework. However, it is often overlooked that achieving superior performance in the simulation experiment is not entirely linearly correlated with achieving better visual effects in the real experiment. Simulation experiments primarily address the losses caused by the reconstruction compression process, whereas real experiments introduce noise that simulates real-world conditions, thus requiring the model to possess stronger generalization capabilities.

However, the degradation caused by insufficient generalization capability is, in fact, not easy to notice. As a result, most current work has not recognized this issue. This oversight is mainly due to the limited test data from real-world experiments, which makes certain issues less likely to be noticed. In most scenes, an improvement in quantitative results in simulation experiments is indeed associated with better performance in real scenes, which has led to some confusion. In this paper, we provide a detailed explanation of this phenomenon, making the problem more apparent. Notably, certain optimization strategies commonly used in previous simulation experiments can even deteriorate specific features in real-world scenes. We assume this phenomenon arises due to the model's overemphasizing of compression loss learning, thereby neglecting other noise (like shot noise) and the semantic information inherent in images. Experimental results in this paper confirm this point.

In this paper, we propose CS-Mamba, a spatial-spectral SSM with our cross-scanning method and local enhancement, to improve HSI reconstruction in SCI. 
Our CS-Mamba aims to strike a balance between global and local information utilization in HSIs. 
CS-Mamba comprises two main components: spatial SSM and spectral-spatial cross SSM. 
The spatial SSM employs a combination of global and local scanning methods, enhancing row-by-row scanning with both global and local modeling. 
Meanwhile, the spectral-spatial cross SSM incorporates an innovative across-scanning mechanism, leveraging spatial-spectral local cubes to enhance the module's ability to capture local details.
Besides, we introduce a masked training strategy in real scenarios. This strategy forces the model to learn the inherent features of images, enhancing the generalization capabilities of previous methods and thereby achieving better results in real scenes.
Overall, our contributions are summarized as follows:
\begin{itemize}
    \item We present a novel spectral-spatial state space model for HSI reconstruction, which can simultaneously integrate information from both spectral and spatial dimensions to adequately model spatial context and spectral features. 
    \item We propose a cross-scanning method that integrates local features from both spatial and spectral dimensions within a single scan. Additionally, we introduce a local scanning approach within the spatial SSM to further enhance local feature extraction.
    \item We introduce a masked training strategy to promote the robustness of models in real scenarios. 
    \vspace{-2mm}
\end{itemize}

\begin{figure}[!t]
    \centering
    \includegraphics[width=0.98\linewidth]{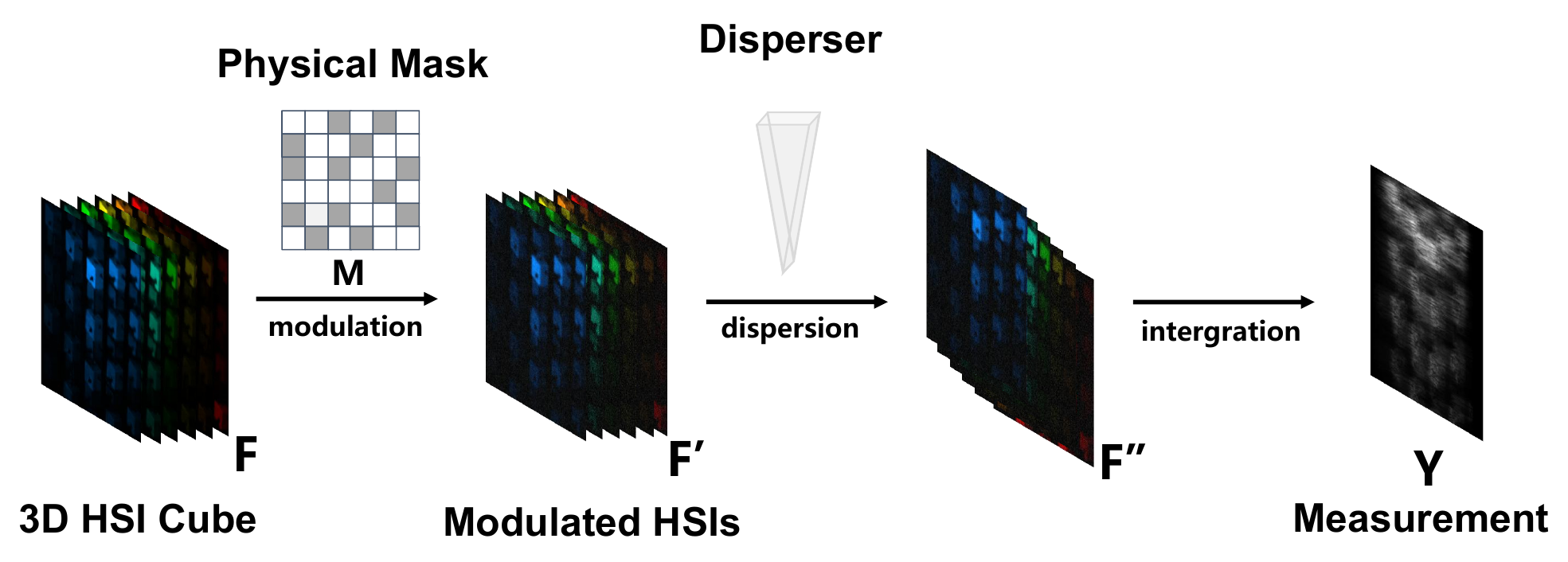} 
    \vspace{-4mm}
    \caption{Schematic of the CASSI system: A 3D HSI cube is modulated by a mask, sheared by a disperser, and transformed into a 2D-coded measurement.}
    \label{fig:CASSI}
    \vspace{-6mm}
\end{figure}

\section{Preliminaries and Related Works}
 \vspace{-1mm} 
\textbf{CASSI System.}
An imaging process CASSI system is illustrated in Fig.~\ref{fig:CASSI}. The 3D HSI data cube is denoted as $\mathbf{F} \in \mathbb{R}^{H \times W \times N_{\lambda} }$ where H, W, and $N_{\lambda}$ represent the height, width, and number of wavelengths of the 3D HSI, respectively. 
A physical mask $\mathbf{M} \in \mathbb{R}^{H \times W}$ is used to modulate the 3D HSI $\mathbf{F}$ as:
      $\mathbf{F^{'}}(:,:,n_{\lambda}) = \mathbf{F^{'}}(:,:,n_{\lambda}) \odot \mathbf{M},$
where $\odot$ represents the element-wise product and $\mathbf{F^{'}} \in \mathbb{R}^{H \times W \times N_{\lambda}} $ is the modulated 3D HSIs. 
After the modulated cube passes the disperser, $\mathbf{F^{'}}$ is tilted and considered to be sheared along the y-axis.
We use $\mathbf{F^{''}} \in \mathbb{R}^{H \times (W+d(N_{\lambda}-1)) \times N_{\lambda}} $ to denote the tilted cube and assume $n_{\lambda}$ to be the reference wavelength. That is, image $\mathbf{F^{'}}(:,:,n_{\lambda_{c}})$ is not sheared along the y-axis. Then we have: 
      $\mathbf{F^{''}}(u,v,n_{\lambda}) = \mathbf{F^{'}} (x,y + d(\lambda_{n} - \lambda_{c}),n_{\lambda})$,
where $(u, v)$ indicates the coordinate system on the detector plane, and $\lambda_{n}$ is the wavelength of channel $n_{\lambda}$. Here,
$d(\lambda_{n} - \lambda_{c})$ signifies the spatial shifting for band $n_{\lambda}$. Finally, the measurement $ \mathbf{Y} \in \mathbb{R}^{H \times [W+d(N_{\lambda}-1)]} $ can be obtained by
    \begin{equation}\label{eqn-3}
      \mathbf{Y} = \sum_{n_{\lambda}=1}^{N_{\lambda}}\mathbf{F^{''}}(:,:,n_{\lambda}) + \mathbf{N},
    \end{equation}
where $ \mathbf{N} \in \mathbb{R}^{H \times [W + d(N_{\lambda}-1)]} $ represents the noise of CASSI caused by the sensing detector.

\noindent \textbf{State Space Models.}
    Recent advancements in state space models (SSMs) have spurred the development of various innovative methods and architectures. Notably, the structured state space model (S4) \cite{s4gu2021} has emerged as a pioneering approach due to its effectiveness in capturing long-range dependencies.
    S4 maps a 1D sequence $x(t) \in \mathbb{R} \rightarrow y(t) \in \mathbb{R}$ by an implicit latent state $h(t)\in \mathbb{R}^{N}$ and decodes it into an output sequence, which can be formulated as:
        \begin{equation}\label{eqn-4}
            \begin{aligned}
            & h^{\prime}(t)=\mathbf{A} h(t)+\mathbf{B} x(t), \\
            & y(t)=\mathbf{C} h(t)+\mathbf{D} x(t),
            \end{aligned}
        \end{equation}
    where $N$ is the state size, ${\rm \textbf{A}} \in \mathbb{R}^{N\times N}$, ${\rm \textbf{B}} \in \mathbb{R}^{N \times 1}$, ${\rm \textbf{C}} \in \mathbb{R}^{1\times N}$, and ${\rm \textbf{D}} \in \mathbb{R}$.
    After that, a discretization process is adopted to integrate Eq. \eqref{eqn-4} into the deep learning algorithm, using the  zero-order hold (ZOH) rule  defined as, 
        \begin{equation}\label{eqn-5}
            \begin{aligned}
                \overline{\rm \textbf{A}} &= {\rm exp}({\rm {\Delta \textbf{A}}}),\\
                \overline{\rm \textbf{B}}&=({\rm {\Delta \textbf{A}}})^{-1}({\rm exp(\textbf{A})}-\textbf{I})\cdot {\rm \Delta \textbf{B}},
            \end{aligned}
        \end{equation}
    where $\overline{\rm \textbf{A}}$ and $\overline{\mathbf{B}}$ represent the discretized forms of the parameters $\mathbf{A}$ and $\mathbf{B}$, respectively. $\rm \Delta$ is denoted as the timescale parameter to transform the continuous parameters $\mathbf{A}$ and $\mathbf{B}$ into discrete ones. 
    

\noindent \textbf{Mamba in Vision.}
Selective structured state space models (S6), or Mamba \cite{mambagu2023mamba}, use a selective mechanism for data-dependent dynamics, enhancing language and visual task performance. Vision Mamba (Vim) \cite{vision_mamba} addresses Mamba's unidirectional modeling by adding bidirectional SSMs for global context and position embeddings for visual recognition. VMamba \cite{liu2024vmamba} bridges 1-D and 2-D scanning via the Cross-Scan Module (CSM). CU-Mamba \cite{cumamba} adds Channel SSM to capture channel correlations, while S$^{2}$Mamba \cite{s2mamba} introduces spectral scanning for semantic retrieval in HSI data. However, these methods overlook local similarity in both spatial and spectral dimensions of HSIs.


\begin{figure*}[t]
	\begin{center}
		\begin{tabular}[t]{c} \hspace{-2.4mm}
			\includegraphics[width=0.98\textwidth]{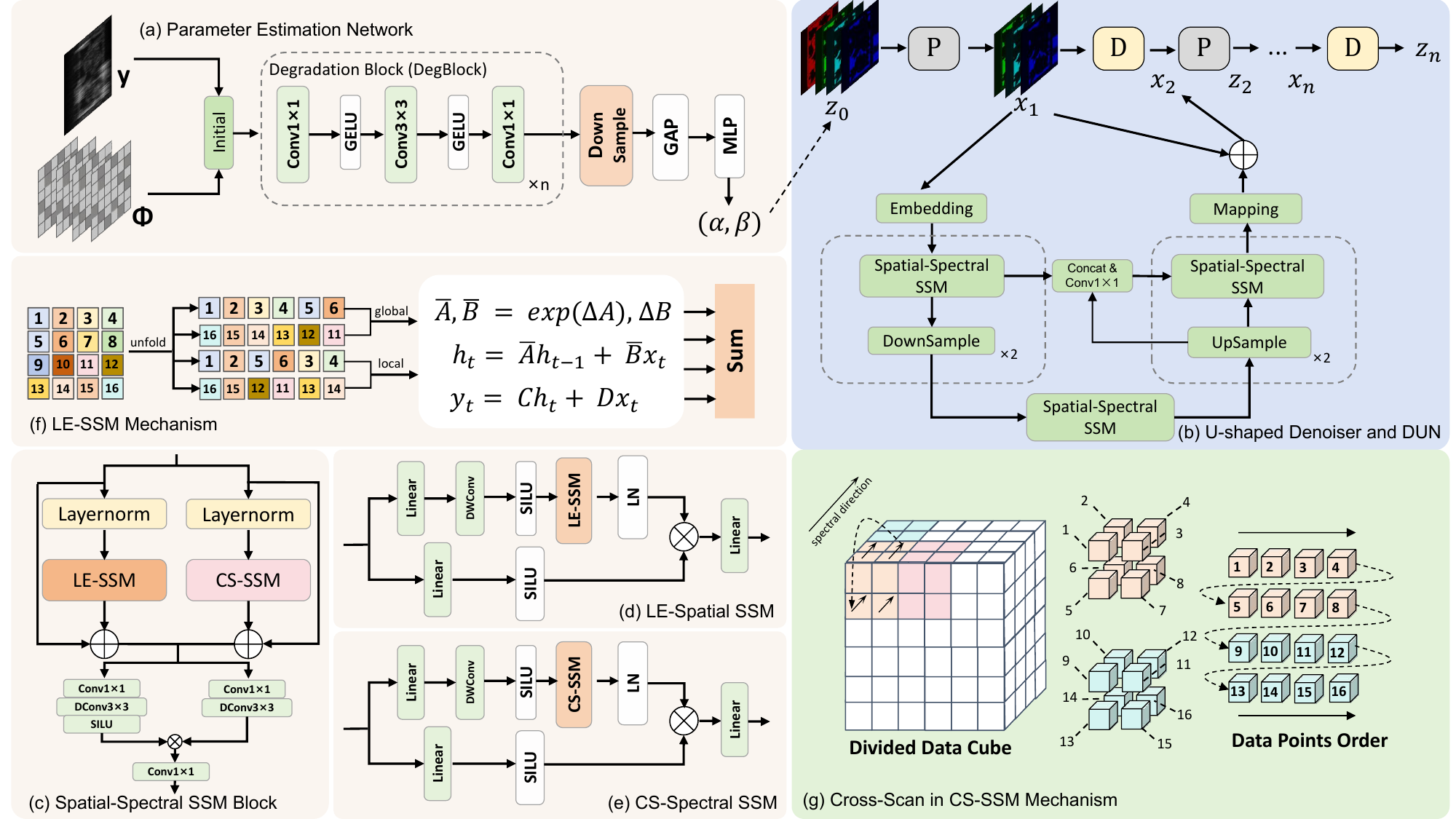}
		\end{tabular}
	\end{center}
	\vspace{-8mm}
	\caption{\small The flowchart of the proposed \textbf{CS-Mamba}. 
    (a) Parameter Estimation Network. This module is used for capturing parameter $\alpha$ and noise level $\beta$ to guide the reconstruction process.
    (b) The U-shaped denoiser network and DUN framework.
    (c) The Spatial-Spectral SSM Block adopted in (b). This module mainly consists of a spatial SSM and a spectral SSM, depicted in the sub-flowchart (d) and (e), respectively.
    (f) The specific method of LE-SSM in the Spatial SSM Module. Scanning directions are divided into global and local branches. 
    (g) The illustration of our Cross-Scan Mechanism. The scanning process is demonstrated by labeling the order of each data point.
    }
	\label{fig:pipeline}
	\vspace{-6mm}
\end{figure*}

\section{The Proposed CS-Mamba Method}
 \vspace{-1mm} 
\subsection{Architecture of CS-Mamba}
 \vspace{-1mm} 
The architecture of our CS-Mamba is illustrated in Fig.~\ref{fig:pipeline}. CS-Mamba leverages a deep unfolding framework guided by mask degradation patterns. Using MAP theory, we model the HSI reconstruction process as follows:
\begin{equation}
\mathbf{\hat{x}} = \text{arg}~\underset{\mathbf{x}}{\text{min}}~~\frac{1}{2} || \mathbf{y} - \mathbf{\Phi} \mathbf{x} ||^2 + \tau R(\mathbf{x}), 
\label{eq:energy_1}
\end{equation}
where $|| \mathbf{y} - \mathbf{\Phi} \mathbf{x} ||^2$ is the data fidelity term, $R(\mathbf{x})$ represents the image prior, and $\tau$ controls the balance between the data fidelity and prior terms. 


HQS introduces an auxiliary variable $z$, resulting in a constrained optimization problem given by
\begin{equation}
    \hat x = \arg \min_{x} \frac{1}{2} \| y - \Phi x \|^2 + \lambda R(z) \quad s.t. \quad z = x.
    \label{eq: aux z}
\end{equation}
Eq. (\ref{eq: aux z}) is solved by minimizing
\begin{equation}
    \mathcal{L}_\mu(x, z) = \frac{1}{2} \| y - \Phi x\|^2 + \lambda R(z) + \frac{\mu}{2}\| z - x \|^2,
    \label{eq: penalty mu}
\end{equation}
where $\mu$ is a penalty parameter. Subsequently, Eq. (\ref{eq: penalty mu}) can be addressed by iteratively solving a data subproblem (x-subproblem) and a prior subproblem (z-subproblem):
\begin{subequations}
    \begin{gather}
        {x_{k} = \arg \min_x \|y - \Phi x\|^2 + \mu \|x - z_{k-1}\|^2,}
        \label{eq: dp} \\
        {z_k = \arg \min_z \frac{1}{2(\sqrt{\lambda / \mu})^2}} \| z - x_k \|^2 + R(z).
        \label{eq: pp}
    \end{gather}
\end{subequations}

The data subproblem, Eq. (\ref{eq: dp}), usually has a fast closed-form solution as
\begin{equation}
    x_k = (\Phi^\top \Phi + \mu I)^{-1}(\Phi^\top y + \mu z_{k-1}),
    \label{eq: close-form}
\end{equation}
where $I$ is an identity matrix, the matrix inversion in Eq. (\ref{eq: close-form}) can be written as
\begin{equation}
    (\Phi^\top \Phi + \mu I)^{-1} = \mu^{-1}I - \mu^{-1}\Phi^\top(I + \Phi \mu^{-1} \Phi^\top)^{-1} \Phi \mu^{-1}.
    \label{eq: partial of close-form}
\end{equation}

In CASSI systems, $\Phi \in \mathbb{R}^{HW' \times HW'N_{\lambda}}$ is a block diagonal matrix, thus, $\Phi \Phi^\top$ is a diagonal matrix that can be defined as 
\begin{equation}
    \Phi \Phi^\top \stackrel{def}{=} \text{diag}\{\phi_1, ..., \phi_i, ..., \phi_{HW'}\}.
    \label{eq: PhiPhiT}
\end{equation}
 By plugging $\Phi \Phi^\top$ into $(I + \Phi \mu^{-1} \Phi^\top)^{-1}$, we obtain:
\begin{equation}
    (I + \Phi \mu^{-1} \Phi^\top)^{-1} = \text{diag}\{\frac{\mu}{\mu + \phi_1}, ..., \frac{\mu}{\mu + \phi_i}, ..., \frac{\mu}{\mu + \phi_{HW'}}\}.
    \label{eq: eq9}
\end{equation}
By plugging Eq. (\ref{eq: partial of close-form}), Eq. (\ref{eq: PhiPhiT}) and Eq. (\ref{eq: eq9}) into Eq. (\ref{eq: close-form}) and simplifying the formula, we have
\begin{equation}
    x_k = z_{k-1} + \Phi^\top [(y - \Phi z_{k-1}) ./ (\mu + diag(\Phi \Phi^\top))],
    \label{eq: close form}
\end{equation}
where $diag$ represents the extraction of elements from its diagonal to form a vector.

The prior subproblem Eq. (\ref{eq: pp}), from a Bayesian perspective, corresponds to Gaussian denoising on $x_k$ with noise level $\sqrt{\lambda/\mu}$. To address this, Eq. (\ref{eq: pp}) could be rewritten as,
\begin{equation}
    z_k = Denoiser(x_k, \sqrt{\lambda/\mu}).
    \label{eq: denoiser}
\end{equation}

DUNs iteratively solve the data subproblem (Eq.~\ref{eq: close form}) and the prior subproblem (Eq. \ref{eq: denoiser}) to address ill-posed inverse problems. The data subproblem depends on the degradation matrix, which can be estimated using either a sensing
matrix \cite{mst,cst} or a DNN \cite{dauhst}. However, sensing
matrices may not accurately reflect real-world degradations, and directly modeling them is challenging. The prior subproblem is treated as a denoising task, but the noise level is unknown in reconstruction, affecting performance. Following \cite{dernn}, we iteratively update the degradation matrix and parameter estimates, learning constant noise and regularization parameters towards each stage due to computational cost, as detailed in the supplementary material.

For the denoiser network, we employ a customized Mamba using a spatial-spectral SSM block as its fundamental unit. The architecture comprises an encoder, a bottleneck, and a decoder. Given a measurement $\mathbf{y}  \in \mathbb{R}^{H(W + d(N_{\lambda} - 1))}$, we reverse the dispersion and shift back $\mathbf{Y}$ to initialize the input signal $\mathbf{X} \in \mathbb{R}^{H \times W \times N_{\lambda} }$:
\begin{equation}\label{eqn-6}
  \mathbf{X}(x,y,n_{\lambda}) = \mathbf{Y} (x,y - d(\lambda_{n} - \lambda_{c})).
\end{equation}
We first concatenate the input $\mathbf{X_{0}}$ with the physical mask $\mathbf{Y} \in \mathbb{R}^{H \times W \times N_{\lambda} }$ and then employ a 1$\times$1 convolution layer to integrate information from the original signal and the physical mask. Moreover, the integrated input $\mathbf{X_{0}}$ undergoes a 3$\times$3 convolution embedding into a feature $\mathbf{X_{0}^{'}} \in \mathbb{R}^{H \times W \times C }$. 
$\mathbf{X_{0}^{'}}$ undergoes downsampling and upsampling before entering the decoder, followed by a 3$\times$3 convolution to extract residual information. Summing the integrated input and residual output yields the reconstructed HSIs. We adjust the number of spatial-spectral SSM blocks in the encoder, bottleneck, and decoder to create different CS-Mamba model sizes.

 \vspace{-1mm} 
\subsection{Enhanced Local SSM (LE-SSM)}
 \vspace{-1mm} 


Although Mamba effectively models long-range dependencies, directly using the SSM mechanism often limits its ability to capture local features. Recent methods integrate CNNs into the network to leverage their local feature extraction capabilities, enhancing original SSM performance. However, this increases the parameter count, undermining the lightweight nature of SSMs. While adding CNN can boost performance, it does not address the core issue.
Inspired by LocalMamba \cite{localmamba}, we propose a simplified local mechanism in the spectral dimension. This involves using two global scanning directions, traversing the image row by row in both forward and backward directions, while the other two scanning directions serve as local branches to unfold and scan local patches.

\begin{figure}[!t]
\centering
\includegraphics[width=0.98\linewidth]{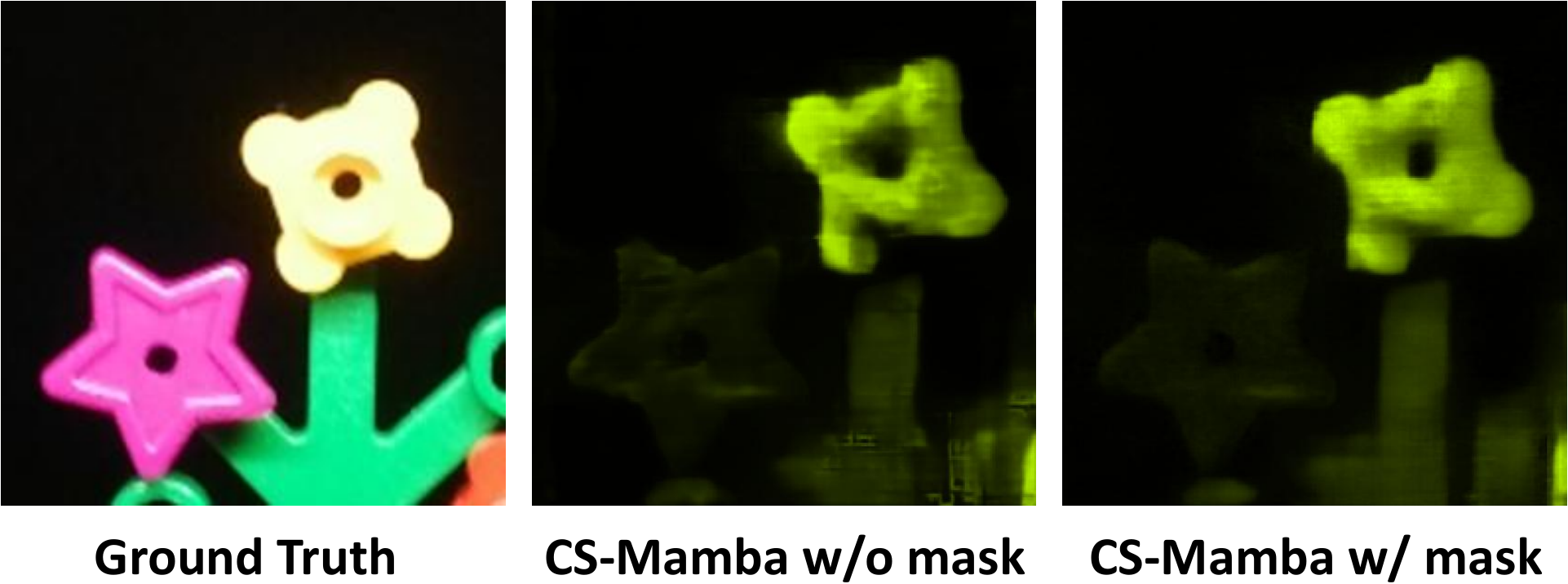} 
\vspace{-3mm}
\caption{A visual comparison of with and w/o masked methods in our CS-Mamba model on real scene 1.}
\label{fig:originmask}
\vspace{-4mm}
\end{figure}

\begin{table}
\centering
    \caption{\footnotesize Comparisons of our denoiser with previous End-to-End networks on PSNR, SSIM, FLOPs, Params, and Inference Time on the test dataset.}
\vspace{-3mm}
\label{tab:denoiser}
 \def\arraystretch{0.9}
\setlength{\tabcolsep}{2.4pt}
	\resizebox{0.45\textwidth}{!}
	{
	\begin{tabular}{l c c c c}
				\toprule
				\rowcolor{color3}Framework &~MST-S~\cite{mst}~ &~CST-S~\cite{cst}~ &~Cross-Scan SSM~   \\
				\midrule
				PSNR &34.14   &34.53   &\bf 34.74  \\
				SSIM  &0.938    &0.945   &\bf0.946  \\
				Params (M)  &0.93    &1.20  &\bf0.81   \\
				FLOPS (G) &12.96    &11.67 &\bf7.01    \\
                    Inference Time(s) &0.022    &0.041 &\bf0.009    \\
				\bottomrule
	\end{tabular}}
	\label{tab:ablations}
 \vspace{-5mm}
\end{table}

    The specific details of the scanning process are illustrated in Fig.~\ref{fig:pipeline}(d). 
    Given the input $\mathbf{X} \in \mathbb{R}^{H \times W \times C }$, we first generate four copies of input data with different arrangements.
    For the global branch, we directly flatten the feature map into $\mathbf{X_{1}} \in \mathbb{R}^{L \times C }$, where $L =  H \times W$, and then reverse it to generate $\mathbf{X_{2}}$. For the local branch, we first set the size of our patch flattening each patch into a 1D sequence, and then linking each patch together to generate $\mathbf{X_{3}}$. 
    Similarly, we reverse $\mathbf{X_{3}}$ to obtain $\mathbf{X_{4}}$. 
    After the generation of input data in four directions, we conduct the process as follows:
        \vspace{-1mm}
        \begin{equation}\label{eqn-7}
            \begin{aligned}
            &\mathbf{X^{'}} = Concat(\mathbf{X_{1}},\mathbf{X_{2}},\mathbf{X_{3}},\mathbf{X_{4}}),\\
            &\mathbf{\hat{X}} = LE\text{-}SSM(\mathbf{X^{'}}).
            \end{aligned}
        \end{equation}

    The structure of LE-SSM is shown in Fig. \ref{fig:pipeline}(b). 
    It is worth noting that our method does not employ any CNNs or spatial and channel attention mechanisms based on MLPs, ensuring that our method does not add any extra parameters or time consumption compared to the global cross-scan mechanism \cite{liu2024vmamba}.
    However, it also means we need to carefully choose the size of local patches, or else the reconstruction result will even become worse. 
    In our implementation, we use patches of size 4$\times$4 to achieve a relatively better performance.

\subsection{Cross-Scanning SSM (CS-SSM) Mechanism}

Our LE-SSM mechanism creates a locally enhanced global receptive field but does not fully utilize the rich spectral information in HSIs. Recent methods \cite{cumamba,s2mamba} have integrated channel information by scanning along the spectral dimension; however, this approach overlooks local features in both spectral and spatial dimensions. Neighboring pixels and bands often share similarities, but direct spectral scanning can increase the distance between highly correlated data points, reducing reconstruction quality.

\begin{table}[H]
\vspace{-3mm}
\centering
    \caption{\footnotesize Comparisons of our CS-Mamba-S (9stg) with previous Transformer-based methods on PSNR, SSIM, FLOPs, and Params.}
\vspace{-2mm}
\label{tab:smallmodel}
 \def\arraystretch{0.9}
\setlength{\tabcolsep}{2.4pt}
	\resizebox{0.42\textwidth}{!}
	{
	\begin{tabular}{l c c c c}
				\toprule
				\rowcolor{color3}Framework &~DAUHST~\cite{dauhst}~ &~DERNN~\cite{dernn}~ &~CS-Mamba-S~   \\
				\midrule
				PSNR &38.36   &39.93   &\bf 40.00  \\
				SSIM  &0.967    &0.976   &\bf0.980  \\
				Params (M)  &6.15M    &0.65M  &\bf0.57M   \\
				FLOPS (G) &79.50    &81.99 &\bf65.77    \\
				\bottomrule
	\end{tabular}}
	\label{tab:ablations}
 \vspace{-5mm}
\end{table}

To address the above issues, we propose a local scanning method that integrates local features from both spatial and spectral dimensions. This mechanism attempts to exploit the local properties of HSIs in both spatial and spectral dimensions while leveraging the rich spectral information to guide the reconstruction process. 
First, we rearrange the input $\mathbf{X} \in \mathbb{R}^{H \times W \times C }$ into patches along the spatial context 
by dividing the input HSIs into a series of smaller data cubes $\mathbf{S} \in \mathbb{R}^{P \times P \times C }$, which will be flattened into sequences for scanning. 
After that, we organize the scanning order within each sequence based on our spatial-spectral local cubes $\mathbf{C} \in \mathbb{R}^{h \times w \times c }$.
As shown in Fig. \ref{fig:pipeline}(e), different colors indicate each spatial-spectral local cube. The scanning order of each data point is also depicted. The process can be formulated as follows:
    \begin{equation}\label{eqn-8}
        \begin{aligned}
        &\mathbf{C} = Divide(\mathbf{X}),\\
        &\mathbf{C^{'}} = CS\text{-}SSM(\mathbf{C}),
        \end{aligned}
    \end{equation}
where $\mathbf{C^{'}}$ denotes the output patch of CS-SSM. After the SSM mechanism, we reorganize the patches back into the same order as the input $\mathbf{X}$. 
By scanning the data points within each data cube, we can leverage the local similarity between adjacent spectral bands and pixels, thus achieving better results than previous methods that directly scan along the spectral dimension.
\vspace{-1mm}

\begin{table*}[t]
	\caption{Comparisons between CS-Mamba and SOTA methods on 10 simulation scenes (S1$\sim$S10). Params, FLOPs, PSNR (upper entry in each cell), and SSIM (lower entry in each cell) are reported. The best and second-best results are highlighted.}
\vspace{-4mm}
 \def\arraystretch{0.9}
\setlength{\tabcolsep}{1.2pt}
	\newcommand{\tabincell}[2]{\begin{tabular}{@{}#1@{}}#2\end{tabular}}
	\centering
	\resizebox{0.99\textwidth}{!}
	{
		\centering
		\begin{tabular}{ccccccccccccccc}
			\bottomrule[0.15em]
			\rowcolor{lightgray}
			~~~~~Algorithms~~~~~
             &~~Reference~~
			&~~Params~~
                &~~GFLOPs~~
			& ~~~~~S1~~~~~
			& ~~~~~S2~~~~~
			& ~~~~~S3~~~~~
			& ~~~~~S4~~~~~
			& ~~~~~S5~~~~~
			& ~~~~~S6~~~~~
			& ~~~~~S7~~~~~
			& ~~~~~S8~~~~~
			& ~~~~~S9~~~~~
			& ~~~~~S10~~~~~
			& ~~~~Avg~~~~
			\\
			\midrule
			DIP-HSI \cite{self}
            & ICCV 2021
			& \tabincell{c}{33.85M}
               & 64.42
			&\tabincell{c}{31.32\\0.855}
			&\tabincell{c}{25.89\\0.699}
			&\tabincell{c}{29.91\\0.839}
			&\tabincell{c}{38.69\\0.926}
			&\tabincell{c}{27.45\\0.796}
            &\tabincell{c}{29.53\\0.824}
			&\tabincell{c}{27.46\\0.700}
            &\tabincell{c}{27.69\\0.802}
			&\tabincell{c}{33.46\\0.863}
			&\tabincell{c}{26.10\\0.733}
			&\tabincell{c}{29.75\\0.803}
			\\
			\midrule
			ADMM-Net \cite{admm-net}
            & ICCV 2019
			& \tabincell{c}{4.27M}
               & 78.58
			&\tabincell{c}{34.03\\0.919}
			&\tabincell{c}{33.57\\0.904}
			&\tabincell{c}{34.82\\0.933}
			&\tabincell{c}{39.46\\0.971}
			&\tabincell{c}{31.83\\0.924}
			&\tabincell{c}{32.47\\0.926}
			&\tabincell{c}{32.01\\0.898}
			&\tabincell{c}{30.49\\0.907}
			&\tabincell{c}{33.38\\0.917}
			&\tabincell{c}{30.55\\0.899}
			&\tabincell{c}{33.26\\0.920}
			\\
                \midrule
			DGSMP \cite{gsm}
            & TPAMI 2023
			& \tabincell{c}{3.76M}
               & 646.65
			&\tabincell{c}{33.26\\0.915}
			&\tabincell{c}{32.09\\0.898}
			&\tabincell{c}{33.06\\0.925}
			&\tabincell{c}{40.54\\0.964}
			&\tabincell{c}{28.86\\0.882}
			&\tabincell{c}{33.08\\0.937}
			&\tabincell{c}{30.74\\0.886}
			&\tabincell{c}{31.55\\0.923}
			&\tabincell{c}{31.66\\0.911}
			&\tabincell{c}{31.44\\0.925}
			&\tabincell{c}{32.63\\0.917}
			\\
                \midrule
                

                 DiffSCI~\cite{diffsci}
                 & CVPR 2024
			& \tabincell{c}{93.52M}
               & 193.76
			&\tabincell{c}{34.96\\0.907}
			&\tabincell{c}{34.60\\0.905}
			&\tabincell{c}{39.83\\0.949}
			&\tabincell{c}{42.65\\0.951}
			&\tabincell{c}{35.21\\0.946}
			&\tabincell{c}{33.12\\0.917}
			&\tabincell{c}{36.29\\0.944}
			&\tabincell{c}{30.42\\0.887}
			&\tabincell{c}{37.27\\0.931}
			&\tabincell{c}{28.49\\0.821}
			&\tabincell{c}{35.28\\0.916}
                \\
                \midrule
                
			MST-L \cite{mst}
            & CVPR 2022
			& \tabincell{c}{2.03M}
               & 28.15
			&\tabincell{c}{35.28\\0.946}
			&\tabincell{c}{35.93\\0.948}
			&\tabincell{c}{36.19\\0.955}
			&\tabincell{c}{41.64\\0.977}
			&\tabincell{c}{32.83\\0.950}
			&\tabincell{c}{34.61\\0.957}
			&\tabincell{c}{33.95\\0.932}
			&\tabincell{c}{32.80\\0.953}
			&\tabincell{c}{34.86\\0.947}
			&\tabincell{c}{32.66\\0.946}
			&\tabincell{c}{35.07\\0.951}
			\\
                \midrule
                
			CST-L+ \cite{cst}
            & ECCV 2022
			& \tabincell{c}{3.00M}
               & 40.10
			&\tabincell{c}{35.64\\ 0.951}
			&\tabincell{c}{36.79\\ 0.957}
			&\tabincell{c}{37.71\\ 0.965}
			&\tabincell{c}{41.38\\ 0.981}
			&\tabincell{c}{32.95\\ 0.957}
			&\tabincell{c}{35.58\\0.966}
			&\tabincell{c}{34.54\\ 0.947}
			&\tabincell{c}{34.07\\ 0.964}
			&\tabincell{c}{35.62\\ 0.959}
			&\tabincell{c}{32.82\\ 0.949}
			&\tabincell{c}{35.71\\ 0.960}
                \\
                \midrule
                
                DAUHST-9stg \cite{dauhst}
                & NIPS 2022
                & 6.15M
                & 79.50
                &\tabincell{c}{37.25\\0.958}
                &\tabincell{c}{39.02\\0.967}
                &\tabincell{c}{41.05\\0.971}
                &\tabincell{c}{46.15\\0.983}
                &\tabincell{c}{35.80\\0.969}
                &\tabincell{c}{37.08\\0.970}
                &\tabincell{c}{37.57\\0.963}
                &\tabincell{c}{35.10\\0.966}
                &\tabincell{c}{40.02\\0.970}
                &\tabincell{c}{34.59\\0.956}
                &\tabincell{c}{38.36\\0.967}
                \\
                \midrule

                PADUT \cite{padut}
                & ICCV 2023
                & 5.38M
                & 90.46
                &\tabincell{c}{37.36\\0.962}
            &\tabincell{c}{40.43\\0.978}
            &\tabincell{c}{42.38\\0.979}
            &\tabincell{c}{46.62\\0.990}
            &\tabincell{c}{36.26\\0.974}
            &\tabincell{c}{37.27\\0.974}
            &\tabincell{c}{37.83\\0.966}
            &\tabincell{c}{35.33\\0.974}
            &\tabincell{c}{40.86\\0.978}
            &\tabincell{c}{34.55\\0.963}
            &\tabincell{c}{38.89\\0.974}
                \\
                \midrule
                
                RDLUF \cite{rdluf}
                & CVPR 2023
                & 1.89M
                & 115.34
                &\tabincell{c}{37.94\\0.966}
            &\tabincell{c}{40.95\\0.977}
            &\tabincell{c}{43.25\\0.979}
            &\tabincell{c}{47.83\\0.990}
            &\tabincell{c}{37.11\\0.976}
            &\tabincell{c}{37.47\\0.975}
            &\tabincell{c}{38.58\\0.969}
            &\tabincell{c}{35.50\\0.970}
            &\tabincell{c}{41.83\\0.978}
            &\tabincell{c}{35.23\\0.962}
            &\tabincell{c}{39.57\\0.974}
                \\
                \midrule

                DERNN (9stg) \cite{dernn}
                & TGRS 2024
                & 0.65M
                & 81.99
                &\tabincell{c}{38.26 \\ 0.965}
                &\tabincell{c}{40.97 \\ 0.979}
                &\tabincell{c}{43.22 \\ 0.979}
                &\tabincell{c}{48.10 \\ 0.991}
                &\tabincell{c}{38.08 \\ 0.980}
                &\tabincell{c}{37.41 \\ 0.975}
                &\tabincell{c}{38.83 \\ 0.971}
                &\tabincell{c}{36.41 \\ 0.973}
                &\tabincell{c}{42.87 \\ 0.981}
                &\tabincell{c}{35.15 \\ 0.962}
                &\tabincell{c}{39.93 \\ 0.976}
                \\
                \midrule

                DERNN (9stg)* \cite{dernn}
                & TGRS 2024
                & 1.04M
                & 134.18
                &\tabincell{c}{38.49 \\ 0.968}
                &\tabincell{c}{41.27 \\ 0.980}
                &\tabincell{c}{\underline{43.97} \\ 0.980}
                &\tabincell{c}{\underline{48.61} \\ 0.992}
                &\tabincell{c}{38.29 \\ 0.981}
                &\tabincell{c}{37.81 \\ 0.977}
                &\tabincell{c}{\textbf{39.30} \\ 0.973}
                &\tabincell{c}{36.51 \\ 0.974}
                &\tabincell{c}{\textbf{43.38} \\ 0.983}
                &\tabincell{c}{35.61\\ 0.966}
                &\tabincell{c}{\underline{40.33} \\ 0.977}
                \\
                \midrule

                SSR-L* \cite{SSR}
                & CVPR 2024
                & 1.73M
                & 78.93
                &\tabincell{c}{\textbf{38.81} \\ 0.968}
                &\tabincell{c}{\underline{41.51} \\ 0.979}
                &\tabincell{c}{43.76 \\ 0.979}
                &\tabincell{c}{\textbf{48.62} \\ 0.988}
                &\tabincell{c}{38.32 \\ 0.979}
                &\tabincell{c}{37.85 \\ 0.975}
                &\tabincell{c}{38.50 \\ 0.969}
                &\tabincell{c}{\underline{36.85} \\ 0.974}
                &\tabincell{c}{42.64 \\ 0.980}
                &\tabincell{c}{\textbf{35.82} \\ 0.965}
                &\tabincell{c}{40.27 \\ 0.976}
                \\
                \midrule
                
                \rowcolor{light-yellow}
			\bf CS-Mamba (3stg) \quad
            & Ours
			& \tabincell{c}{0.97M}
               & 37.80
			&\tabincell{c}{37.93\\0.968}
			&\tabincell{c}{40.00\\0.978}
			&\tabincell{c}{42.73\\0.983}
			&\tabincell{c}{47.18\\0.993}
			&\tabincell{c}{36.48\\0.976}
			&\tabincell{c}{37.10\\0.976}
			&\tabincell{c}{37.91\\0.970}
			&\tabincell{c}{35.39\\0.974}
			&\tabincell{c}{41.43\\0.981}
			&\tabincell{c}{34.82\\0.965}
			&\tabincell{c}{39.10\\0.977}
			\\
                \midrule
                \rowcolor{light-yellow}
			\bf CS-Mamba (5stg)
            & Ours
			& \tabincell{c}{0.97M}
               & 62.93
			&\tabincell{c}{38.23\\0.971}
			&\tabincell{c}{40.62\\0.982}
			&\tabincell{c}{43.44\\0.984}
			&\tabincell{c}{47.58\\0.994}
			&\tabincell{c}{37.95\\0.982}
			&\tabincell{c}{37.74\\0.979}
			&\tabincell{c}{38.71\\0.974}
			&\tabincell{c}{36.27\\0.979}
			&\tabincell{c}{42.89\\0.986}
			&\tabincell{c}{35.45\\\underline{0.970}}
			&\tabincell{c}{39.90\\0.980}
			\\
                \midrule
                \rowcolor{light-yellow}
			\bf CS-Mamba (9stg)
            & Ours 
			& \tabincell{c}{0.97M}
               & 113.18
			&\tabincell{c}{38.51\\0.973}
			&\tabincell{c}{41.30\\\underline{0.984}}
			&\tabincell{c}{43.77\\0.985}
			&\tabincell{c}{47.55\\0.994}
			&\tabincell{c}{\underline{38.72}\\\underline{0.985}}
			&\tabincell{c}{\underline{38.05}\\\underline{0.981}}
			&\tabincell{c}{38.88\\0.976}
			&\tabincell{c}{36.64\\\underline{0.981}}
			&\tabincell{c}{43.06\\\underline{0.987}}
			&\tabincell{c}{35.54\\0.971}
			&\tabincell{c}{40.21\\ \underline{0.982}}

                \\
                \midrule
                \rowcolor{light-yellow}
                \bf CS-Mamba* (9stg)
            & Ours
			& \tabincell{c}{1.16M}
               & 124.98
			&\tabincell{c}{\underline{38.68}\\\textbf{0.974}}
			&\tabincell{c}{\textbf{41.55}\\\textbf{0.985}}
			&\tabincell{c}{\textbf{43.99}\\\textbf{0.986}}
			&\tabincell{c}{47.86\\\textbf{0.994}}
			&\tabincell{c}{\textbf{38.85}\\\textbf{0.986}}
			&\tabincell{c}{\textbf{38.15}\\\textbf{0.982}}
			&\tabincell{c}{\underline{39.09}\\\textbf{0.976}}
			&\tabincell{c}{\textbf{36.86}\\\textbf{0.983}}
			&\tabincell{c}{\underline{43.25}\\\textbf{0.988}}
			&\tabincell{c}{\underline{35.78}\\\textbf{0.971}}
			&\tabincell{c}{\textbf{40.42}\\ \textbf{0.983}}
			\\
                \midrule

                \rowcolor{light-yellow}
                \bf CS-Mamba-s (9stg)
            & Ours
			& \tabincell{c}{\textbf{0.57M}}
               & 65.77
			&\tabincell{c}{38.63\\0.972}
			&\tabincell{c}{40.90\\0.983}
			&\tabincell{c}{43.46\\0.984}
			&\tabincell{c}{47.30\\0.994}
			&\tabincell{c}{38.41\\0.983}
			&\tabincell{c}{37.68\\0.979}
			&\tabincell{c}{38.80\\\underline{0.975}}
			&\tabincell{c}{36.29\\0.979}
			&\tabincell{c}{43.09\\0.986}
			&\tabincell{c}{35.41\\0.969}
			&\tabincell{c}{40.00\\ 0.980}
			\\
                \midrule

			\toprule[0.15em]
		\end{tabular}
	}
	\vspace{-6mm}
	\label{tab:simu}
\end{table*}

    \begin{figure}[!t]
    \centering
    \includegraphics[width=0.98\linewidth]{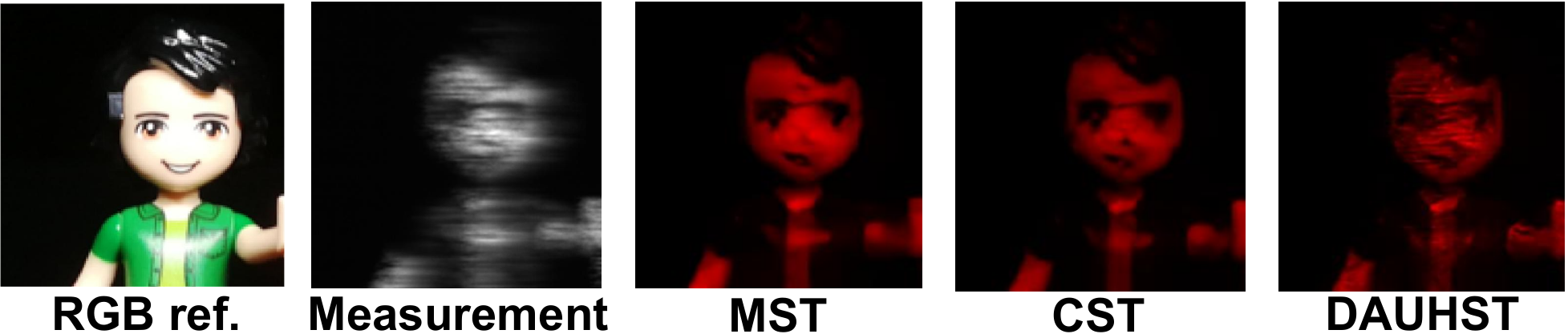} 
    \vspace{-3mm}
    \caption{A visual comparison of the performance of MST, CST, and DAUHST methods in the real scene.}
    \label{fig:mstcstdauhst}
    \vspace{-6mm}
    \end{figure}

\subsection{Masked Training Strategy}

Before presenting our masked training strategy, we discuss our observed phenomena and motivation for this approach.
In Fig.~\ref{fig:mstcstdauhst}, we display visual results for three Transformer-based methods \cite{mst,cst,dauhst} in a real scene. While their simulation performance improves from left to right, visual quality degrades in real scenarios. Compared to MST, CST's reconstructed image shows more noise artifacts on the nose and forehead. These spots, absent in ground truth, suggest noise from 2D compressed measurements. DAUHST’s results exhibit even greater degradation, which is consistently observed across other algorithms in this scene.

This effect stems from two factors. First, simulation-based image reconstruction, aimed at minimizing compression loss, struggles with unknown real-world noise distributions. While shot noise is added during training, it inadequately reflects real-world complexities, leading to insufficient generalization. Secondly, focusing on mask pattern recovery can overemphasize specific information, weakening resilience to other noise types, and creating a balance challenge between compression and real-world losses.
    \begin{equation}\label{eqn-9}
        \begin{aligned}
        &Total\ Loss = Compression\ Loss + \epsilon,
        \end{aligned}
    \end{equation}  
where $\epsilon$ represents the complex real noise. 
We propose a masked training strategy inspired by \cite{masktrain} for SCI reconstruction to enhance robustness and generalization. During training, a randomly generated 0-1 mask is applied to feature maps, creating an inpainting task that compels the network to reconstruct occluded regions. This strengthens the model’s capacity to handle complex real-world scenarios. Unlike traditional image denoising, where noise is added to clean images, SCI reconstruction starts with compressed 2D measurements that inherently carry compression loss, compounded by mixed noise in practice. By consistently applying the same mask during both training and testing, we ensure alignment between phases, leading to improved reconstruction accuracy and visually appealing results. This approach addresses unique SCI challenges, enhancing model performance under varied real-world conditions.

\vspace{-2mm} 
\section{Experiment}
\vspace{-1mm}
\subsection{Experimental Settings}
\vspace{-1mm}
The effectiveness of our CS-Mamba network is verified on both simulation and real datasets, by using 28 wavelengths ranging from 450nm to 650nm, acquired via spectral interpolation manipulation.

\noindent \textbf{Simulation HSI Data.}
Two simulated HSI datasets CAVE \cite{cave} and KAIST \cite{kaist} are utilized to conduct the simulation experiment. 
The CAVE dataset contains 32 HSIs with a spatial size of $512\times512$, while the KAIST dataset includes 30 HSIs with a spatial size of $2704\times 3376$. 
Following the experimental settings of TSA-Net \cite{tsa_net}, we designate CAVE as the training set and select 10 scenes from the KAIST as the testing set.

\noindent \textbf{Real HSI Data.} 
We adopt the real HSIs obtained by the CASSI system, as developed in TSA-Net \cite{tsa_net} for testing.

\noindent \textbf{Evaluation Metrics.} 
We select peak signal-to-noise ratio (PSNR) \cite{PSNR} and structural similarity (SSIM) \cite{ssim} as the evaluation metrics to assess the quality of HSI reconstruction.

\noindent \textbf{Implementation Details.} 
Our CS-Mamba is implemented using PyTorch and we conduct model training with Adam optimizer ($\beta_{1}$ = 0.9 and $\beta_{2}$ = 0.999) for 300 epochs.
The learning rate is set to $1\times10^{-3}$ at the beginning and a cosine annealing
scheduler is adopted. Image patches with size $256\times256$ are randomly cropped from the 3D HSI cubes with 28 channels for training in the simulation and real experiments. 
The shifting step $d$ in the dispersion process is set to 2.
We apply the data augmentation techniques including random flipping and rotation. Models are trained on RTX 4090 GPUs.

\begin{figure*}[!t]
	\centering
	\renewcommand{\h}{0.105}
	\renewcommand{\wa}{0.12}
	\newcommand{\wb}{0.16}
	\renewcommand{\g}{-0.7mm}
	\renewcommand{\tabcolsep}{1.8pt}
	\renewcommand{\arraystretch}{1}

        \resizebox{1\linewidth}{!} {
		\begin{tabular}{cc}			
			\renewcommand{\name}{figs/simulation/}
			\renewcommand{\h}{0.5}
			\renewcommand{\w}{0.5}
			\begin{tabular}{cc}
                    \Huge
				\begin{adjustbox}{valign=t}
					\begin{tabular}{cc}%
		         	\includegraphics[trim={0 0 0 0 },clip, width=1.09\textwidth]{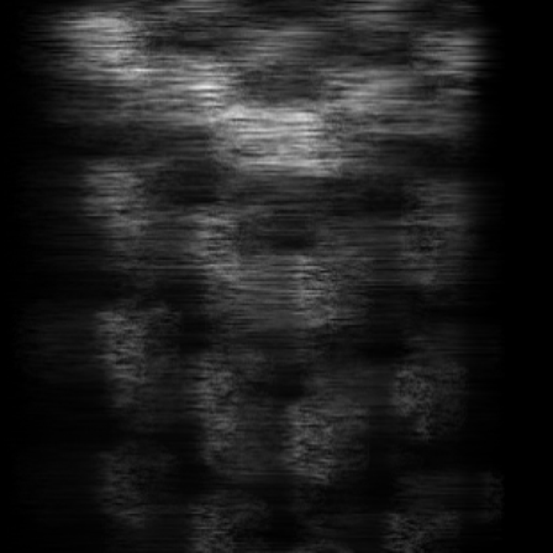}
						\\
						\resizebox{0.3\textwidth}{!}{Scene 2}
					\end{tabular}
				\end{adjustbox}
				\begin{adjustbox}{valign=t}
					\begin{tabular}{ccccccc} 

                        \includegraphics[trim={50 40 150 160 },clip,height=\h \textwidth, width=\w \textwidth]{\name measurement/meas2.png} \hspace{\g} &
                        \includegraphics[trim={50 40 150 160 }, clip,height=\h \textwidth, width=\w \textwidth]{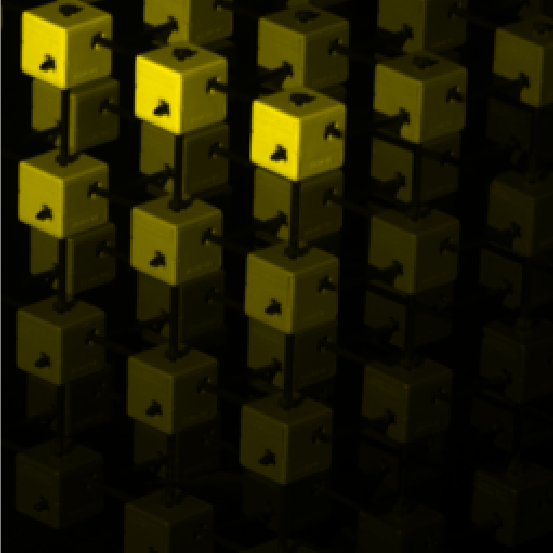} \hspace{\g} &
                        \includegraphics[trim={50 40 150 160 }, clip,height=\h \textwidth, width=\w \textwidth]{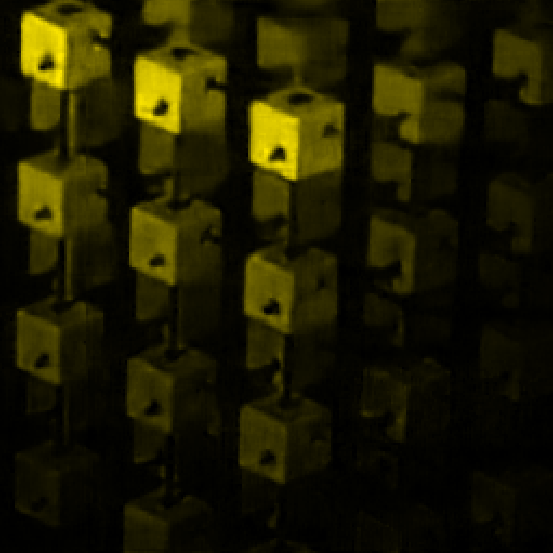} \hspace{\g} &
						\includegraphics[trim={50 40 150 160 }, clip,height=\h \textwidth, width=\w \textwidth]{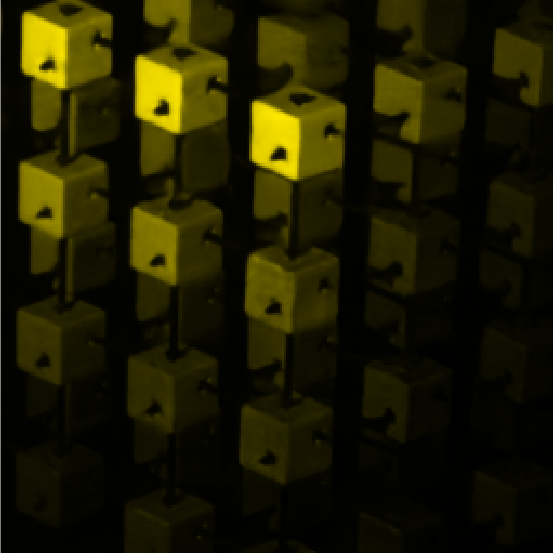} \hspace{\g} &
      					\includegraphics[trim={50 40 150 160 }, clip,height=\h \textwidth, width=\w \textwidth]{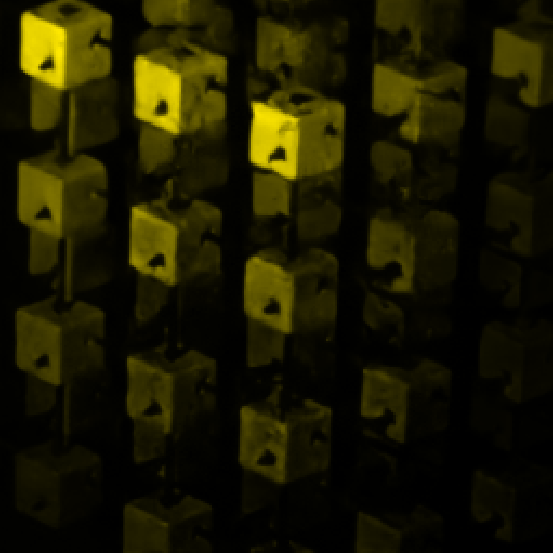} \hspace{\g} &
						\includegraphics[trim={50 40 150 160 }, clip,height=\h \textwidth, width=\w \textwidth]{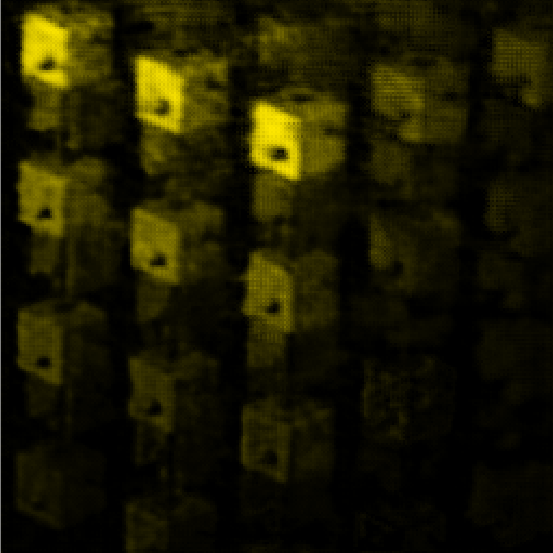} \hspace{\g} &
 						\includegraphics[trim={50 40 150 160 }, clip,height=\h \textwidth, width=\w \textwidth]{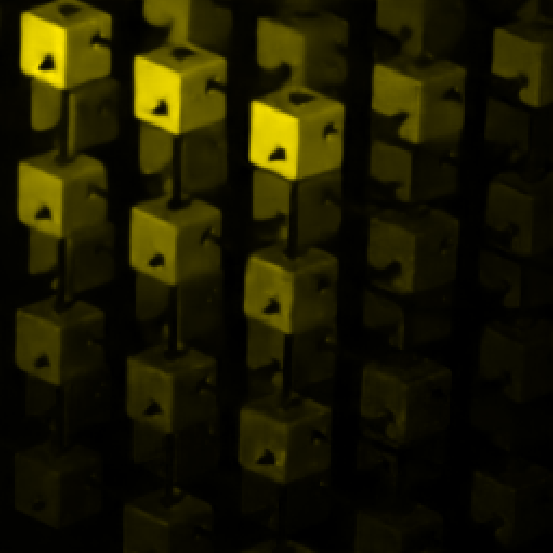} \hspace{\g} 
					  \\
                        \resizebox{0.22\textwidth}{!}{Meas } &\resizebox{0.32\textwidth}{!}{GT Patch } & \resizebox{0.4\textwidth}{!}{ADMM-Net } & \resizebox{0.25\textwidth}{!}{CST-L+} & \resizebox{0.3\textwidth}{!}{DGSMP } & \resizebox{0.3\textwidth}{!}{DIP-HSI}  & \resizebox{0.25\textwidth}{!}{HDNet } 
                            \\
      						
					   \includegraphics[trim= {50 40 150 160 }, clip,height=\h \textwidth, width=\w \textwidth]{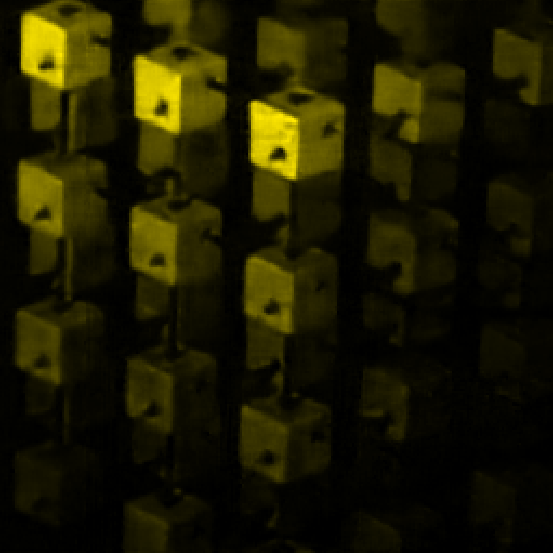} \hspace{\g} &	\includegraphics[trim={50 40 150 160 }, clip,height=\h \textwidth, width=\w \textwidth]{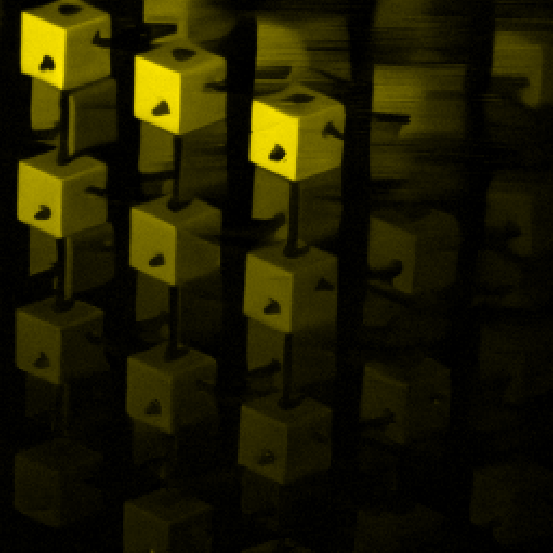} \hspace{\g} &
						\includegraphics[trim={50 40 150 160 }, clip,height=\h \textwidth, width=\w \textwidth]{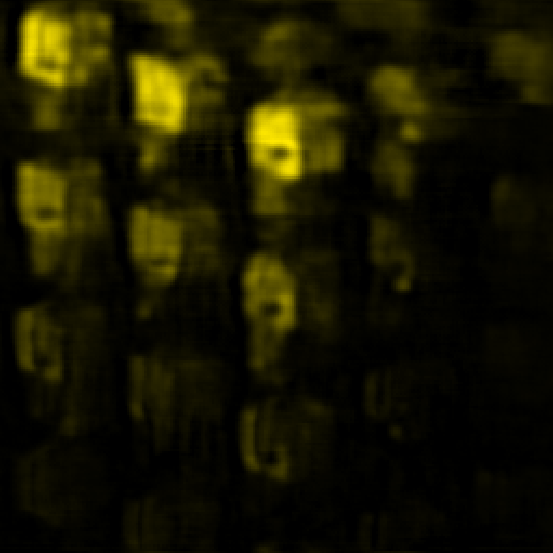} \hspace{\g} &
      					\includegraphics[trim={50 40 150 160 }, clip,height=\h \textwidth, width=\w \textwidth]{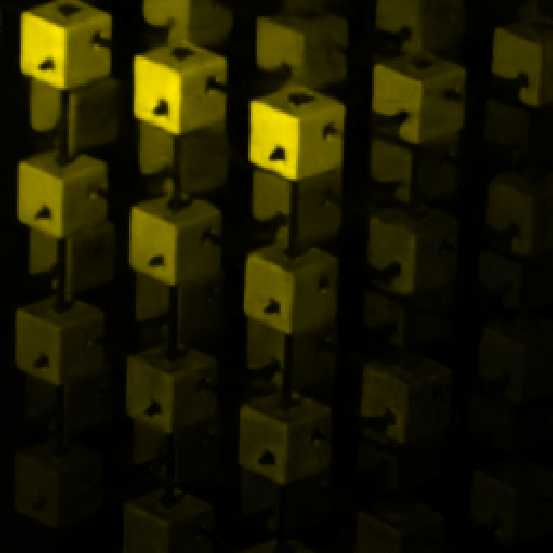} \hspace{\g} &

                        \includegraphics[trim={50 40 150 160 }, clip,height=\h \textwidth, width=\w \textwidth]{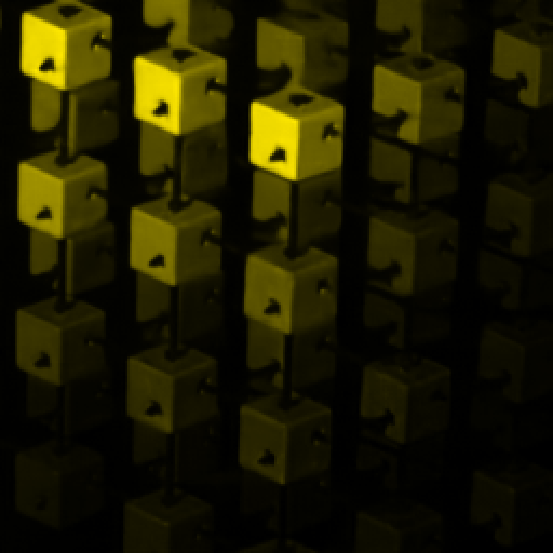} \hspace{\g} &	
						\includegraphics[trim={50 40 150 160 }, clip,height=\h \textwidth, width=\w \textwidth]{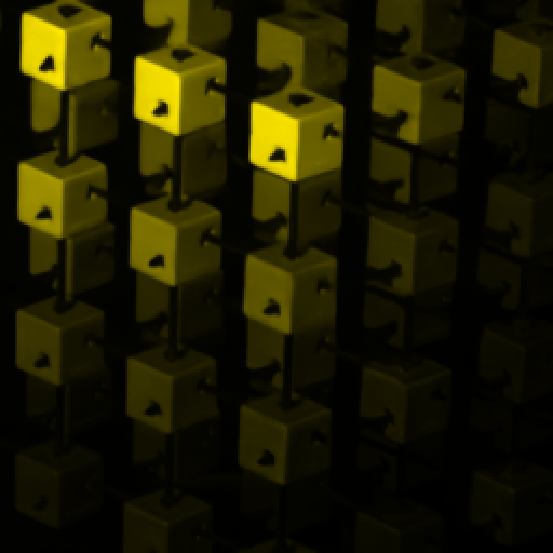} \hspace{\g} &		
						\includegraphics[trim={50 40 150 160 }, clip,height=\h \textwidth, width=\w \textwidth]{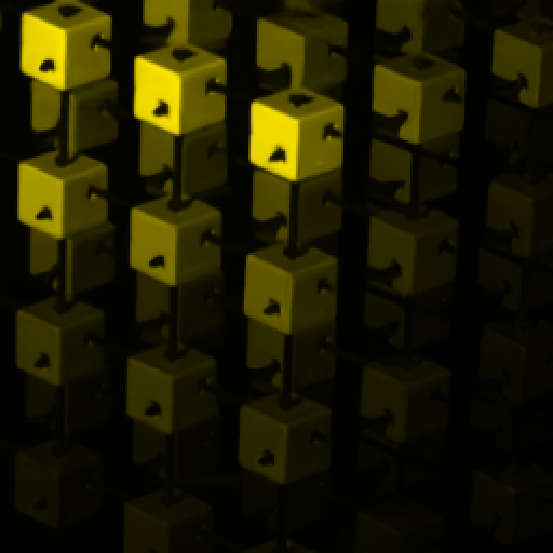} \hspace{\g} \\
                        \resizebox{0.33\textwidth}{!}{GAP-Net }&
                        \resizebox{0.3\textwidth}{!}{DiffSCI } & \resizebox{0.25\textwidth}{!}{$\lambda$-Net } & \resizebox{0.25\textwidth}{!}{MST-L} & \resizebox{0.4\textwidth}{!}{DAUHST} & \resizebox{0.35\textwidth}{!}{PADUT } &
                        \resizebox{0.45\textwidth}{!}{\textbf{CS-Mamba} }
					\end{tabular}
				\end{adjustbox}
			\end{tabular}	
		\end{tabular}
}
	\vspace{-4mm}
	\caption{Visual comparison of different methods on simulation 
    $Scene$ 2 at wavelength 575.5nm} 
    \vspace{-5mm}
	\label{fig_kaist_s1}
\end{figure*}

\vspace{-2mm}
\subsection{Simulation Results}
\vspace{-2mm}

We conduct a comprehensive comparative analysis of our CS-Mamba method and SOTA HSI reconstruction algorithms in Params, FLOPs, PSNR, and SSIM. 
We compared the results of CS-Mamba with 11 SOTA methods algorithms across 10 simulation scenes. The quantitative results of different models are listed in Table~\ref{tab:simu}.

As shown in Table~\ref{tab:simu}, our CS-Mamba-9stg* obtains 40.42dB of PSNR and 0.983 of SSIM, achieving the best reconstruction performance among SOTAs.
When compared to recent Transformer-based methods such as DAUHST \cite{dauhst}, RDLUF \cite{rdluf}, DERNN \cite{dernn}, and SSR-L* \cite{SSR}, our CS-Mamba demonstrates notable improvements over these methods, which are 2.06dB, 0.85dB, 0.49dB, and 0.15dB on average, respectively. 

We also conduct comparisons of our CS-Mamba-S with previous Transformer-based methods DAUHST \cite{dauhst} and DERNN \cite{dernn}.
As shown in Table~\ref{tab:smallmodel}, we not only achieve better performance in PSNR and SSIM compared to DAUHST and DERNN but also obtain a more lightweight model. Specifically, the CS-Mamba has 0.08 (M) fewer parameters than the DERNN and 13.73 (G) fewer in FLOPs, which means the smallest parameter count and Computational cost among recent deep unfolding methods. Furthermore, we evaluated the denoiser network (Cross-Scan SSM Block) as an independent end-to-end model, comparing it against previous end-to-end Transformer methods, as presented in Table~\ref{tab:denoiser}. 
Specifically, the Spatial-Spectral SSM is 2.4 times faster than MST-S and 4.5 times faster than CST-S, while having 0.12 (M) fewer parameters than the MST-S, which means the smallest parameter count.

Except for quantitative comparisons, we also visualize the results of simulation experiments. 
We choose one of 28 channels from Scene 2, against the simulation results obtained from eleven SOTA approaches. 
Visually, as illustrated in Fig.~\ref{fig_kaist_s1}, our method yields smoother and cleaner details. 
The experimental results indicate the capability of our method to generate high-quality HSIs.

\vspace{-1.5mm}
\subsection{Real Data Results}
\vspace{-1mm}

\begin{figure*}[t]
        \vspace{-3mm}
	\centering
	\renewcommand{\h}{0.105}
	\renewcommand{\wa}{0.12}
	\newcommand{\wb}{0.16}
	\renewcommand{\g}{-0.7mm}
	\renewcommand{\tabcolsep}{1.8pt}
	\renewcommand{\arraystretch}{1}
        \resizebox{1\linewidth}{!} {
		\begin{tabular}{cc}			
			\renewcommand{\name}{figs/real/}
			\renewcommand{\h}{0.5}
			\renewcommand{\w}{0.5}
			\begin{tabular}{cc}
                    \Huge
				\begin{adjustbox}{valign=t}
					\begin{tabular}{cc}%
		         	\includegraphics[trim={0 0 0 0 },clip, width=0.75\textwidth]{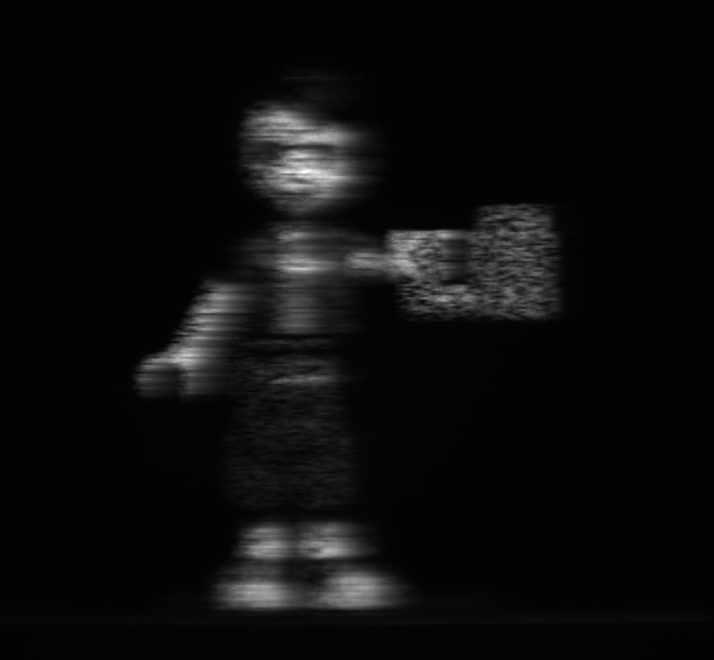}
						\\
						\resizebox{0.55\textwidth}{!}{Real Measurement}
                            \\
                            \includegraphics[trim={0 0 0 0 },clip, width=0.75\textwidth]{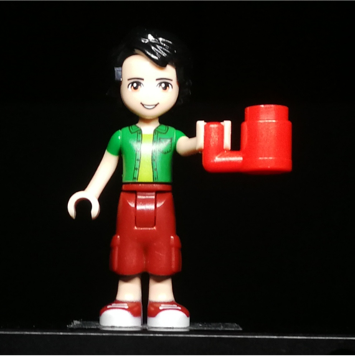}
                            \\
						\resizebox{0.55\textwidth}{!}{Ground Truth}
					\end{tabular}
				\end{adjustbox}
				\begin{adjustbox}{valign=t}
					\begin{tabular}{cccccc}
                        \includegraphics[clip,height=\h \textwidth, width=\w \textwidth]{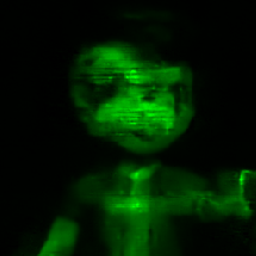} \hspace{\g} & 	
						\includegraphics[clip,height=\h \textwidth, width=\w \textwidth]{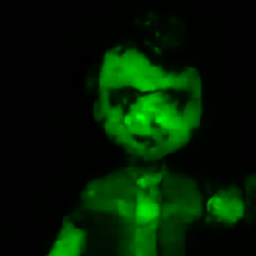} \hspace{\g} &	
						\includegraphics[clip,height=\h \textwidth, width=\w \textwidth]{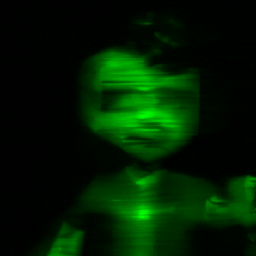} \hspace{\g} &
      					\includegraphics[clip,height=\h \textwidth, width=\w \textwidth]{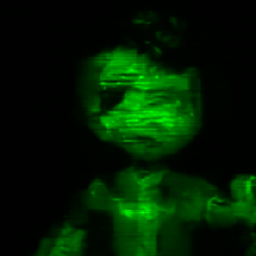} \hspace{\g} &	
						\includegraphics[clip,height=\h \textwidth, width=\w \textwidth]{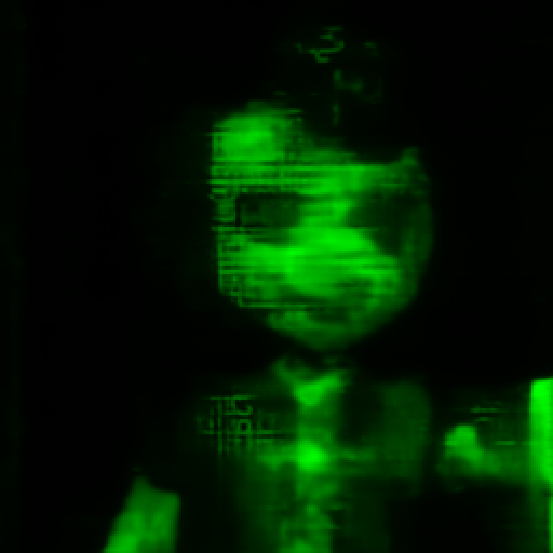} \hspace{\g} &		
						\includegraphics[clip,height=\h \textwidth, width=\w \textwidth]{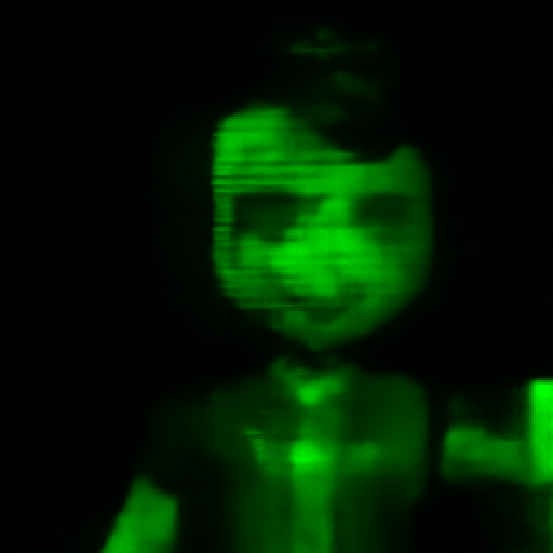}
                        \hspace{\g}

                        \\
      					\includegraphics[clip,height=\h \textwidth, width=\w \textwidth]{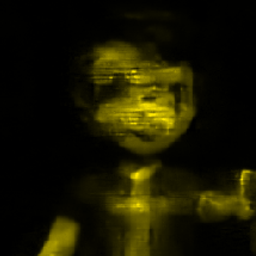} \hspace{\g} & 	
						\includegraphics[clip,height=\h \textwidth, width=\w \textwidth]{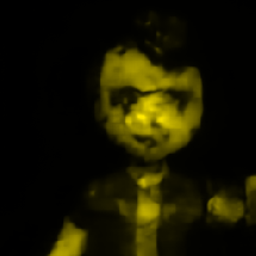} \hspace{\g} &	
						\includegraphics[clip,height=\h \textwidth, width=\w \textwidth]{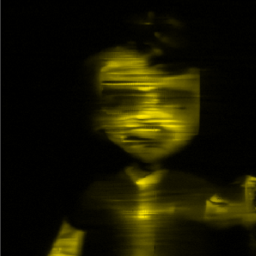} \hspace{\g} &
      					\includegraphics[clip,height=\h \textwidth, width=\w \textwidth]{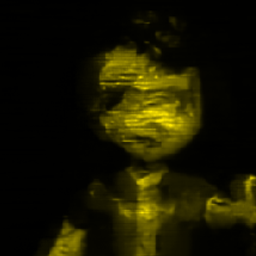} \hspace{\g} &		 
						\includegraphics[clip,height=\h \textwidth, width=\w \textwidth]{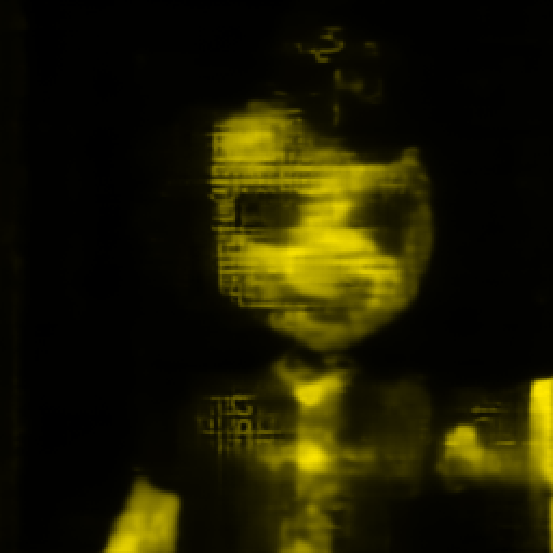} \hspace{\g} &		
						\includegraphics[clip,height=\h \textwidth, width=\w \textwidth]{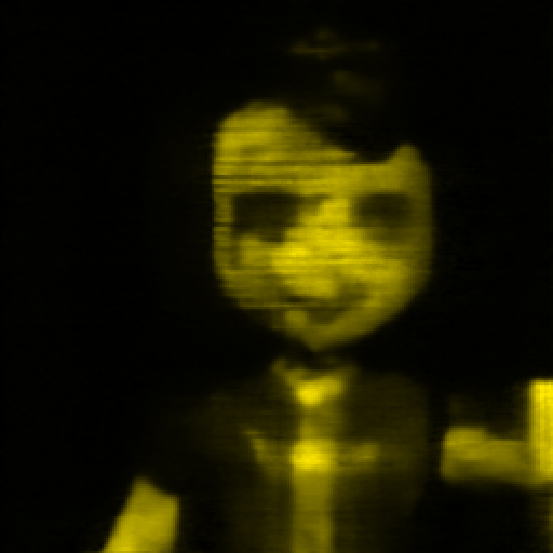}
                        \hspace{\g}

                         \\
      					\includegraphics[clip,height=\h \textwidth, width=\w \textwidth]{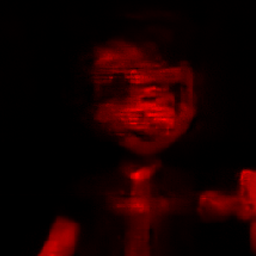} \hspace{\g} & 	
						\includegraphics[clip,height=\h \textwidth, width=\w \textwidth]{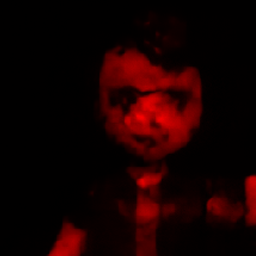} \hspace{\g} &	
						\includegraphics[clip,height=\h \textwidth, width=\w \textwidth]{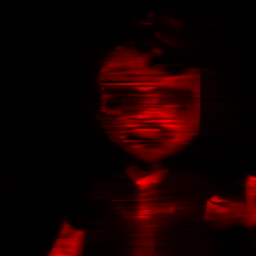} \hspace{\g} &
      					\includegraphics[clip,height=\h \textwidth, width=\w \textwidth]{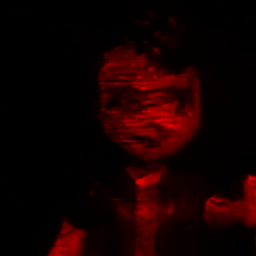} \hspace{\g} &	  
						\includegraphics[clip,height=\h \textwidth, width=\w \textwidth]{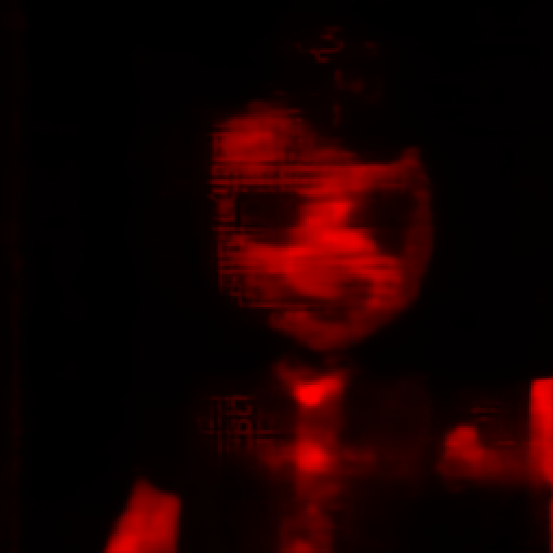} \hspace{\g} &		
						\includegraphics[clip,height=\h \textwidth, width=\w \textwidth]{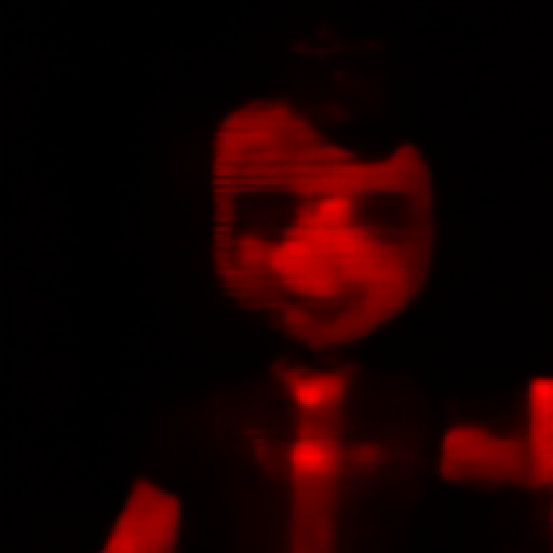}
                        \hspace{\g} \\
                        \resizebox{0.27\textwidth}{!}{GAP-Net} & \resizebox{0.27\textwidth}{!}{DGSMP } & \resizebox{0.32\textwidth}{!}{DiffSCI } & \resizebox{0.27\textwidth}{!}{DAUHST} & \resizebox{0.37\textwidth}{!}{\textbf{Ours (3stg)}} & \resizebox{0.42\textwidth}{!}{\textbf{Ours (Mask)} }
                        
					\end{tabular}
				\end{adjustbox}
			\end{tabular}	
		\end{tabular}
  
	}
        \vspace{-4mm}
	\caption{Visual comparison of SCI reconstruction methods on real $Scene$ 3 at wavelength 648.0nm.} %
	\label{fig_real_s1}
    \vspace{-6mm}
\end{figure*}

We apply our model and the proposed masked strategy to reconstruct real HSIs, retraining on the CAVE \cite{cave} and KAIST \cite{kaist} datasets and simulating real conditions with 11-bit shot noise. Fig.~\ref{fig_real_s1} compares reconstructions from two real scenes using our original CS-Mamba model, the masked version, and four state-of-the-art methods. 
When enlarging the image results, we can clearly observe that the original CS-Mamba output exhibits a peculiar grid-like noise around the shoulders and face of the subject. In contrast, our mask method effectively mitigates this issue, producing a more visually pleasing effect compared to previous approaches, achieving smoother and more accurate reconstruction. 

We further conducted a separate comparison between the original model and the model employing the masked strategy in Scene 1. As shown in Fig.~\ref{fig:originmask}, the original model exhibits some wrinkles and artifacts on the raised areas of the petals, with edges becoming blurred, and the stem displaying certain noise patterns. In contrast, the ground truth image presents smooth elevations. This issue is significantly alleviated with our masked method, obtaining a marked improvement in visual quality.
Additional experiments will be included in the supplementary material to further substantiate the validity of our approach and conclusions, while Visual comparisons here sufficiently demonstrate the effectiveness of our masked training method.

\vspace{-2mm}
\subsection{Ablation Study}
\vspace{-2mm}
\begin{wraptable}{r}{0.249\textwidth}
\vspace{-8mm}
\centering
    \caption{\footnotesize Ablation study of CS-Mamba (3stg)'s components}
\vspace{-2mm}
\label{tab:breakdown}
 \def\arraystretch{1}
\setlength{\tabcolsep}{2.4pt}
	\resizebox{0.25\textwidth}{!}
	{
	 \begin{tabular}{c|c|cc}
                    \toprule
                \rowcolor{color3} Architecture & Params (M) & PSNR & SSIM  \\
                    \midrule
                
                Baseline w/o SSM 
                &     0.37       &   37.41   &      0.964  \\
                    \midrule
                with LE-SSM                                &  0.77 & 38.83 & 0.975    \\
                with CS-SSM                                 &    0.57       & 37.88   &  0.968  \\
                    \midrule
                with LE + CS SSM                     &    0.97       &   \textbf{39.10}   &   \textbf{0.977}    \\
                    \bottomrule
                \end{tabular}}
 \vspace{-5mm}
\end{wraptable}
\textbf{Break-down Ablation}. 
We conduct an ablation study by removing the LE-SSM block and CS-SSM block from the CS-Mamba network. The experiment results are shown in Table~\ref{tab:breakdown}. Our goal is to explore the contributions of each mechanism to the results. The baseline barely adopts the parameter estimation, U-shaped framework, and GDFFN block without any SSM modules. Since \cite{dernn} has demonstrated the effectiveness of GDFFN block, we don't conduct an ablation study for it. 

When we respectively employ LE-SSM and CS-SSM blocks, our model achieves 1.42 dB and 0.47 dB improvements in PSNR over the baseline, respectively. Employing both mechanisms simultaneously can further enhance the model's performance. Here we need to point out that, in consideration of maintaining lightweight, we only utilized a single-direction scan in CS-SSM. In fact, we could also deploy a four-direction scanning similar to LE-SSM. However, each scan direction implies additional computational costs so we decided not to adopt this way.

\begin{wraptable}{r}{0.22\textwidth}
\vspace{-4.5mm}
\centering
    \caption{\footnotesize Comparisons of CS mechanism and common scanning.}
\vspace{-2mm}
\label{tab:AS}
 \def\arraystretch{1}
\setlength{\tabcolsep}{2.4pt}
	\resizebox{0.22\textwidth}{!}
	{
	 \begin{tabular}{c|c|cc}
                    \toprule
                \rowcolor{color3} Architecture                &  PSNR & SSIM  \\
                    \midrule
                
                w/o Spectral SSM 
                &  38.83   &      0.975  \\
                    \midrule
                with CS Mechanism                                   &   \textbf{39.10}   &   \textbf{0.977}    \\
                with Common Scanning                            &  38.95   &   0.975    \\
                    \bottomrule
                \end{tabular}}
 \vspace{-5mm}
\end{wraptable}
\noindent \textbf{Cross-Scan Mechanism}. 
To highlight the advantages of our cross-scan mechanism over common spectral scanning, we compared model performance under both approaches in our CS-Mamba (3stg) network. As shown in Table~\ref{tab:AS}, common scanning provided a 0.12 dB improvement, whereas the proposed across-scanning mechanism yielded a significant 0.27 dB increase. Both methods were tested under single-scan conditions, with our method excelling due to its enhanced focus on adjacent bands and pixels, effectively capturing local features across spatial and spectral dimensions. These results underscore the effectiveness of our across-scanning approach.

\begin{wraptable}{r}{0.22\textwidth}
\vspace{-2mm}
\centering
    \caption{\footnotesize Comparisons of LE mechanism and global scanning.}
\vspace{-2mm}
\label{tab:LE}
 \def\arraystretch{1}
\setlength{\tabcolsep}{2.4pt}
	\resizebox{0.22\textwidth}{!}
	{
	 \begin{tabular}{c|c|cc}
                    \toprule
                \rowcolor{color3} Architecture                &  PSNR & SSIM  \\
                    \midrule
                
                w/o Spatial SSM 
                &  37.88  &      0.968  \\
                    \midrule
                with LE Machenism                                   &   \textbf{38.83}   &   \textbf{0.975}    \\
                with Global Scanning                            &  38.66   &   0.974    \\
                    \bottomrule
                \end{tabular}}
 \vspace{-4mm}
\end{wraptable}
\noindent \textbf{Local-Enhancement Mechanism}. 
We also conducted an ablation study of our LE-SSM mechanism in our CS-Mamba (3stg) to assess its impact. As shown in Table~\ref{tab:LE}, starting from a baseline model with a PSNR of 37.88 dB (without spatial SSM), incorporating global four-direction scanning improved PSNR by 0.95 dB and SSIM by 0.007, demonstrating effective global spatial context modeling. Applying our LE mechanism along the spatial dimension further improved PSNR by 0.17 dB compared to global scanning, without introducing additional parameters or computational costs, unlike methods using CNN layers or spatial attention \cite{localmamba}. This enhancement highlights the significance of capturing local features in HSIs and validates the effectiveness of our LE mechanism.



\vspace{-2mm}

\section{Conclusion}
\vspace{-2mm}
In this paper, we propose a novel cross-scan SSM mechanism that more fully leverages the spectral and spatial local features of HSI images, achieving improved performance compared to the common SSM mechanism. Additionally, we identify and elaborate on noise issues encountered in real-world scene reconstruction, presenting a masked training method designed to address these challenges. This approach significantly mitigates the degradation problems in real-scene reconstruction that previous methods failed to resolve, enhancing the robustness of the model. Experiments in the main text and supplementary material validate the effectiveness of our method.


{
    \small
    \bibliographystyle{ieeenat_fullname}
    \bibliography{main}

\begin{thebibliography}{54}
\providecommand{\natexlab}[1]{#1}
\providecommand{\url}[1]{\texttt{#1}}
\expandafter\ifx\csname urlstyle\endcsname\relax
  \providecommand{\doi}[1]{doi: #1}\else
  \providecommand{\doi}{doi: \begingroup \urlstyle{rm}\Url}\fi

\bibitem[Backman et~al.(2000)Backman, Wallace, Perelman, Arendt, Gurjar, M{\"u}ller, Zhang, Zonios, Kline, McGillican, et~al.]{mi_1}
V Backman, Michael~B Wallace, LT Perelman, JT Arendt, R Gurjar, MG M{\"u}ller, Q Zhang, G Zonios, E Kline, T McGillican, et~al.
\newblock Detection of preinvasive cancer cells.
\newblock \emph{Nature}, 406\penalty0 (6791):\penalty0 35--36, 2000.

\bibitem[Cai et~al.(2022{\natexlab{a}})Cai, Lin, Hu, Wang, Yuan, Zhang, Timofte, and Van~Gool]{cst}
Yuanhao Cai, Jing Lin, Xiaowan Hu, Haoqian Wang, Xin Yuan, Yulun Zhang, Radu Timofte, and Luc Van~Gool.
\newblock Coarse-to-fine sparse transformer for hyperspectral image reconstruction.
\newblock In \emph{Proceedings of the European Conference on Computer Vision}, pages 686--704. Springer, 2022{\natexlab{a}}.

\bibitem[Cai et~al.(2022{\natexlab{b}})Cai, Lin, Hu, Wang, Yuan, Zhang, Timofte, and Van~Gool]{mst}
Yuanhao Cai, Jing Lin, Xiaowan Hu, Haoqian Wang, Xin Yuan, Yulun Zhang, Radu Timofte, and Luc Van~Gool.
\newblock Mask-guided spectral-wise transformer for efficient hyperspectral image reconstruction.
\newblock In \emph{Proceedings of the IEEE/CVF Conference on Computer Vision and Pattern Recognition}, pages 17502--17511, 2022{\natexlab{b}}.

\bibitem[Cai et~al.(2022{\natexlab{c}})Cai, Lin, Lin, Wang, Zhang, Pfister, Timofte, and Van~Gool]{mst_pp}
Yuanhao Cai, Jing Lin, Zudi Lin, Haoqian Wang, Yulun Zhang, Hanspeter Pfister, Radu Timofte, and Luc Van~Gool.
\newblock Mst++: Multi-stage spectral-wise transformer for efficient spectral reconstruction.
\newblock In \emph{Proceedings of the IEEE/CVF Conference on Computer Vision and Pattern Recognition}, pages 745--755, 2022{\natexlab{c}}.

\bibitem[Cai et~al.(2022{\natexlab{d}})Cai, Lin, Wang, Yuan, Ding, Zhang, Timofte, and Gool]{dauhst}
Yuanhao Cai, Jing Lin, Haoqian Wang, Xin Yuan, Henghui Ding, Yulun Zhang, Radu Timofte, and Luc~V Gool.
\newblock Degradation-aware unfolding half-shuffle transformer for spectral compressive imaging.
\newblock In \emph{Proceedings of the Advances in Neural Information Processing Systems}, pages 37749--37761, 2022{\natexlab{d}}.

\bibitem[Cao et~al.(2016)Cao, Yue, Lin, Lin, Yuan, Dai, Carin, and Brady]{sci_5}
Xun Cao, Tao Yue, Xing Lin, Stephen Lin, Xin Yuan, Qionghai Dai, Lawrence Carin, and David~J Brady.
\newblock Computational snapshot multispectral cameras: Toward dynamic capture of the spectral world.
\newblock \emph{IEEE Signal Processing Magazine}, 33\penalty0 (5):\penalty0 95--108, 2016.

\bibitem[Chan and Whiteman(1983)]{PSNR}
Luen~C. Chan and Peter Whiteman.
\newblock Hardware-constrained hybrid coding of video imagery.
\newblock \emph{IEEE Transactions on Aerospace and Electronic Systems}, AES-19\penalty0 (1):\penalty0 71--84, 1983.

\bibitem[Chen et~al.(2023)Chen, Gu, Liu, Magid, Dong, Wang, Pfister, and Zhu]{masktrain}
Haoyu Chen, Jinjin Gu, Yihao Liu, Salma~Abdel Magid, Chao Dong, Qiong Wang, Hanspeter Pfister, and Lei Zhu.
\newblock Masked image training for generalizable deep image denoising.
\newblock In \emph{Proceedings of the IEEE/CVF Conference on Computer Vision and Pattern Recognition (CVPR)}, pages 1692--1703, 2023.

\bibitem[Chen et~al.(2017)Chen, Guo, Wang, Wang, Peng, and He]{chen2017denoising}
Yongyong Chen, Yanwen Guo, Yongli Wang, Dong Wang, Chong Peng, and Guoping He.
\newblock Denoising of hyperspectral images using nonconvex low rank matrix approximation.
\newblock \emph{IEEE Transactions on Geoscience and Remote Sensing}, 55\penalty0 (9):\penalty0 5366--5380, 2017.

\bibitem[Choi et~al.(2017)Choi, Kim, Gutierrez, Jeon, and Nam]{kaist}
Inchang Choi, MH Kim, D Gutierrez, DS Jeon, and G Nam.
\newblock High-quality hyperspectral reconstruction using a spectral prior.
\newblock In \emph{Technical report}, 2017.

\bibitem[Deng~R(2024)]{cumamba}
Gu~T. Deng~R.
\newblock Cu-mamba: Selective state space models with channel learning for image restoration.
\newblock \emph{arXiv preprint arXiv:2404.11778}, 2024.

\bibitem[Dong et~al.(2021)Dong, Zhou, Wu, Wu, Shi, and Li]{dong2021model}
Weisheng Dong, Chen Zhou, Fangfang Wu, Jinjian Wu, Guangming Shi, and Xin Li.
\newblock Model-guided deep hyperspectral image super-resolution.
\newblock \emph{IEEE Transactions on Image Processing}, 30:\penalty0 5754--5768, 2021.

\bibitem[Dong et~al.(2023)Dong, Gao, Qiu, Li, Yang, and Shi]{rdluf}
Yubo Dong, Dahua Gao, Tian Qiu, Yuyan Li, Minxi Yang, and Guangming Shi.
\newblock Residual degradation learning unfolding framework with mixing priors across spectral and spatial for compressive spectral imaging.
\newblock In \emph{Proceedings of the IEEE/CVF Conference on Computer Vision and Pattern Recognition}, pages 22262--22271, 2023.

\bibitem[Dong et~al.(2024)Dong, Gao, Li, Shi, and Liu]{dernn}
Yubo Dong, Dahua Gao, Yuyan Li, Guangming Shi, and Danhua Liu.
\newblock Degradation estimation recurrent neural network with local and non-local priors for compressive spectral imaging.
\newblock \emph{IEEE Transactions on Geoscience and Remote Sensing}, 62:\penalty0 1--15, 2024.

\bibitem[Du et~al.(2009)Du, Tong, Cao, and Lin]{sci_6}
Hao Du, Xin Tong, Xun Cao, and Stephen Lin.
\newblock A prism-based system for multispectral video acquisition.
\newblock In \emph{2009 IEEE 12th International Conference on Computer Vision}, pages 175--182. IEEE, 2009.

\bibitem[Gu and Dao(2023{\natexlab{a}})]{gu2023mamba}
Albert Gu and Tri Dao.
\newblock Mamba: Linear-time sequence modeling with selective state spaces.
\newblock \emph{arXiv preprint arXiv:2312.00752}, 2023{\natexlab{a}}.

\bibitem[Gu and Dao(2023{\natexlab{b}})]{mambagu2023mamba}
Albert Gu and Tri Dao.
\newblock Mamba: Linear-time sequence modeling with selective state spaces.
\newblock \emph{arXiv preprint arXiv:2312.00752}, 2023{\natexlab{b}}.

\bibitem[Gu et~al.(2021)Gu, Goel, and Re]{s4gu2021}
Albert Gu, Karan Goel, and Christopher Re.
\newblock Efficiently modeling long sequences with structured state spaces.
\newblock In \emph{ICLR}, 2021.

\bibitem[Guo et~al.(2024)Guo, Li, Dai, Ouyang, Ren, and Xia]{mambair}
Hang Guo, Jinmin Li, Tao Dai, Zhihao Ouyang, Xudong Ren, and Shu-Tao Xia.
\newblock Mambair: A simple baseline for image restoration with state-space model.
\newblock \emph{arXiv preprint arXiv:2402.15648}, 2024.

\bibitem[Han and Chen(2019)]{sr_3}
Xian-Hua Han and Yen-Wei Chen.
\newblock Deep residual network of spectral and spatial fusion for hyperspectral image super-resolution.
\newblock In \emph{2019 IEEE Fifth International Conference on Multimedia Big Data (BigMM)}, pages 266--270. IEEE, 2019.

\bibitem[He et~al.(2020)He, Yao, Li, Yokoya, Zhao, Zhang, and Zhang]{he2020non}
Wei He, Quanming Yao, Chao Li, Naoto Yokoya, Qibin Zhao, Hongyan Zhang, and Liangpei Zhang.
\newblock Non-local meets global: An iterative paradigm for hyperspectral image restoration.
\newblock \emph{IEEE Transactions on Pattern Analysis and Machine Intelligence}, 44\penalty0 (4):\penalty0 2089--2107, 2020.

\bibitem[Huang et~al.(2024{\natexlab{a}})Huang, Chen, and He]{huang2024spectral}
Lingbo Huang, Yushi Chen, and Xin He.
\newblock Spectral-spatial mamba for hyperspectral image classification.
\newblock \emph{arXiv preprint arXiv:2404.18401}, 2024{\natexlab{a}}.

\bibitem[Huang et~al.(2021)Huang, Dong, Yuan, Wu, and Shi]{gsm}
Tao Huang, Weisheng Dong, Xin Yuan, Jinjian Wu, and Guangming Shi.
\newblock Deep gaussian scale mixture prior for spectral compressive imaging.
\newblock In \emph{Proceedings of the IEEE/CVF Conference on Computer Vision and Pattern Recognition}, pages 16216--16225, 2021.

\bibitem[Huang et~al.(2024{\natexlab{b}})Huang, Pei, You, Wang, Qian, and Xu]{localmamba}
Tao Huang, Xiaohuan Pei, Shan You, Fei Wang, Chen Qian, and Chang Xu.
\newblock Localmamba: Visual state space model with windowed selective scan.
\newblock \emph{arXiv preprint arXiv:2403.09338}, 2024{\natexlab{b}}.

\bibitem[Kim et~al.(2012)Kim, Harvey, Kittle, Rushmeier, J.~Dorsey, and Brady]{ot_1}
M.~H. Kim, T.~A. Harvey, D.~S. Kittle, H. Rushmeier, R.~O.~Prum J.~Dorsey, and D.~J. Brady.
\newblock 3d imaging spectroscopy for measuring hyperspectral patterns on solid objects.
\newblock \emph{ACM Transactions on on Graphics}, 2012.

\bibitem[Li et~al.(2023)Li, Fu, Liu, and Zhang]{padut}
Miaoyu Li, Ying Fu, Ji Liu, and Yulun Zhang.
\newblock Pixel adaptive deep unfolding transformer for hyperspectral image reconstruction.
\newblock In \emph{Proceedings of the IEEE/CVF International Conference on Computer Vision}, pages 12959--12968, 2023.

\bibitem[Liao et~al.(2024)Liao, Zhu, Wang, Pan, Wang, and Ma]{lightm-unet}
Weibin Liao, Yinghao Zhu, Xinyuan Wang, Cehngwei Pan, Yasha Wang, and Liantao Ma.
\newblock Lightm-unet: Mamba assists in lightweight unet for medical image segmentation.
\newblock \emph{arXiv preprint arXiv:2403.05246}, 2024.

\bibitem[Liu et~al.(2024)Liu, Tian, Zhao, Yu, Xie, Wang, Ye, and Liu]{liu2024vmamba}
Yue Liu, Yunjie Tian, Yuzhong Zhao, Hongtian Yu, Lingxi Xie, Yaowei Wang, Qixiang Ye, and Yunfan Liu.
\newblock Vmamba: Visual state space model.
\newblock \emph{arXiv preprint arXiv:2401.10166}, 2024.

\bibitem[Llull et~al.(2013)Llull, Liao, Yuan, Yang, Kittle, Carin, Sapiro, and Brady]{sci_1}
Patrick Llull, Xuejun Liao, Xin Yuan, Jianbo Yang, David Kittle, Lawrence Carin, Guillermo Sapiro, and David~J Brady.
\newblock Coded aperture compressive temporal imaging.
\newblock \emph{Optics Express}, 2013.

\bibitem[Lu and Fei(2014)]{mi_2}
Guolan Lu and Baowei Fei.
\newblock Medical hyperspectral imaging: a review.
\newblock \emph{Journal of Biomedical Optics}, 2014.

\bibitem[Ma et~al.(2019)Ma, Liu, Shou, and Yuan]{admm-net}
Jiawei Ma, Xiao-Yang Liu, Zheng Shou, and Xin Yuan.
\newblock Deep tensor admm-net for snapshot compressive imaging.
\newblock In \emph{Proceedings of the IEEE/CVF International Conference on Computer Vision}, pages 10223--10232, 2019.

\bibitem[Meng et~al.(2020{\natexlab{a}})Meng, Ma, and Yuan]{tsa_net}
Ziyi Meng, Jiawei Ma, and Xin Yuan.
\newblock End-to-end low cost compressive spectral imaging with spatial-spectral self-attention.
\newblock In \emph{Proceedings of the European Conference on Computer Vision}, pages 187--204. Springer, 2020{\natexlab{a}}.

\bibitem[Meng et~al.(2020{\natexlab{b}})Meng, Qiao, Ma, Yu, Xu, and Yuan]{mi_3}
Ziyi Meng, Mu Qiao, Jiawei Ma, Zhenming Yu, Kun Xu, and Xin Yuan.
\newblock Snapshot multispectral endomicroscopy.
\newblock \emph{Optics Letters}, 2020{\natexlab{b}}.

\bibitem[Meng et~al.(2021)Meng, Yu, Xu, and Yuan]{self}
Ziyi Meng, Zhenming Yu, Kun Xu, and Xin Yuan.
\newblock Self-supervised neural networks for spectral snapshot compressive imaging.
\newblock In \emph{Proceedings of the IEEE/CVF International Conference on Computer Vision}, pages 2622--2631, 2021.

\bibitem[Miao et~al.(2019)Miao, Yuan, Pu, and Athitsos]{lambda}
Xin Miao, Xin Yuan, Yunchen Pu, and Vassilis Athitsos.
\newblock $\lambda$-net: Reconstruct hyperspectral images from a snapshot measurement.
\newblock In \emph{Proceedings of the IEEE/CVF International Conference on Computer Vision}, pages 4059--4069, 2019.

\bibitem[Pan et~al.(2003)Pan, Healey, Prasad, and Tromberg]{ot_2}
Zhihong Pan, Glenn Healey, Manish Prasad, and Bruce Tromberg.
\newblock Face recognition in hyperspectral images.
\newblock \emph{IEEE Transactions on Pattern Analysis and Machine Intelligence}, 25\penalty0 (12):\penalty0 1552--1560, 2003.

\bibitem[Pan et~al.(2024)Pan, Zeng, Cao, Zhang, and Chen]{diffsci}
Zhenghao Pan, Haijin Zeng, Jiezhang Cao, Kai Zhang, and Yongyong Chen.
\newblock Diffsci: Zero-shot snapshot compressive imaging via iterative spectral diffusion model.
\newblock In \emph{Proceedings of the IEEE/CVF Conference on Computer Vision and Pattern Recognition}, 2024.

\bibitem[Park et~al.(2007)Park, Lee, Grossberg, and Nayar]{cave}
Jong-Il Park, Moon-Hyun Lee, Michael~D Grossberg, and Shree~K Nayar.
\newblock Multispectral imaging using multiplexed illumination.
\newblock In \emph{2007 IEEE 11th International Conference on Computer Vision}, pages 1--8. IEEE, 2007.

\bibitem[Shi et~al.(2024)Shi, Xia, Jin, Wang, Zhao, Xia, Xiao, and Yang]{shi2024vmambair}
Yuan Shi, Bin Xia, Xiaoyu Jin, Xing Wang, Tianyu Zhao, Xin Xia, Xuefeng Xiao, and Wenming Yang.
\newblock Vmambair: Visual state space model for image restoration.
\newblock \emph{arXiv preprint arXiv:2403.11423}, 2024.

\bibitem[Van~Nguyen et~al.(2010)Van~Nguyen, Banerjee, and Chellappa]{ot_3}
Hien Van~Nguyen, Amit Banerjee, and Rama Chellappa.
\newblock Tracking via object reflectance using a hyperspectral video camera.
\newblock In \emph{IEEE Computer Society Conference on Computer Vision and Pattern Recognition-Workshops}, pages 44--51. IEEE, 2010.

\bibitem[Wagadarikar et~al.(2008)Wagadarikar, John, Willett, and Brady]{sci_2}
Ashwin Wagadarikar, Renu John, Rebecca Willett, and David Brady.
\newblock Single disperser design for coded aperture snapshot spectral imaging.
\newblock \emph{Applied Optics}, 2008.

\bibitem[Wagadarikar et~al.(2009)Wagadarikar, Pitsianis, Sun, and Brady]{sci_3}
Ashwin~A Wagadarikar, Nikos~P Pitsianis, Xiaobai Sun, and David~J Brady.
\newblock Video rate spectral imaging using a coded aperture snapshot spectral imager.
\newblock \emph{Optics Express}, 2009.

\bibitem[Wang et~al.(2024{\natexlab{a}})Wang, Zhang, Peng, Zhang, Jia, and Jiao]{s2mamba}
Guanchun Wang, Xiangrong Zhang, Zelin Peng, Tianyang Zhang, Xiuping Jia, and Licheng Jiao.
\newblock S2mamba: A spatial-spectral state space model for hyperspectral image classification.
\newblock \emph{arXiv preprint arXiv:2404.18213}, 2024{\natexlab{a}}.

\bibitem[Wang et~al.(2023)Wang, Wang, Chen, Hu, Song, and Huang]{sr_2}
Hongyuan Wang, Lizhi Wang, Chang Chen, Xue Hu, Fenglong Song, and Hua Huang.
\newblock Learning spectral-wise correlation for spectral super-resolution: Where similarity meets particularity.
\newblock In \emph{Proceedings of the 31st ACM International Conference on Multimedia}, pages 7676--7685, 2023.

\bibitem[Wang et~al.(2024{\natexlab{b}})Wang, Wang, Ding, Li, Wu, Rong, Kong, Huang, Li, Yang, et~al.]{mamba_survey}
Xiao Wang, Shiao Wang, Yuhe Ding, Yuehang Li, Wentao Wu, Yao Rong, Weizhe Kong, Ju Huang, Shihao Li, Haoxiang Yang, et~al.
\newblock State space model for new-generation network alternative to transformers: A survey.
\newblock \emph{arXiv preprint arXiv:2404.09516}, 2024{\natexlab{b}}.

\bibitem[Wang et~al.(2004)Wang, Bovik, Sheikh, and Simoncelli]{ssim}
Zhou Wang, Alan~C Bovik, Hamid~R Sheikh, and Eero~P Simoncelli.
\newblock Image quality assessment: from error visibility to structural similarity.
\newblock \emph{IEEE Transactions on Image Processing}, 13\penalty0 (4):\penalty0 600--612, 2004.

\bibitem[Yang et~al.(2023)Yang, Xiang, Tong, Zhao, and Li]{icme_sci_2}
Shumian Yang, Xinxin Xiang, Fenghua Tong, Dawei Zhao, and Xin Li.
\newblock Image compressed sensing using multi-scale characteristic residual learning.
\newblock In \emph{2023 IEEE International Conference on Multimedia and Expo (ICME)}, pages 1595--1600. IEEE, 2023.

\bibitem[Ying et~al.(2023)Ying, Wang, Shi, and Yin]{icme_sci_1}
Yangke Ying, Jin Wang, Yunhui Shi, and Baocai Yin.
\newblock Dual-domain feature learning and memory-enhanced unfolding network for spectral compressive imaging.
\newblock In \emph{2023 IEEE International Conference on Multimedia and Expo (ICME)}, pages 1589--1594. IEEE, 2023.

\bibitem[Yuan et~al.(2018)Yuan, Zhang, Li, Shen, and Zhang]{yuan2018hyperspectral}
Qiangqiang Yuan, Qiang Zhang, Jie Li, Huanfeng Shen, and Liangpei Zhang.
\newblock Hyperspectral image denoising employing a spatial--spectral deep residual convolutional neural network.
\newblock \emph{IEEE Transactions on Geoscience and Remote Sensing}, 57\penalty0 (2):\penalty0 1205--1218, 2018.

\bibitem[Yuan(2016)]{gap_tv}
Xin Yuan.
\newblock Generalized alternating projection based total variation minimization for compressive sensing.
\newblock In \emph{2016 IEEE International conference on image processing (ICIP)}, pages 2539--2543. IEEE, 2016.

\bibitem[Yuan et~al.(2015)Yuan, Tsai, Zhu, Llull, Brady, and Carin]{sci_4}
Xin Yuan, Tsung-Han Tsai, Ruoyu Zhu, Patrick Llull, David Brady, and Lawrence Carin.
\newblock Compressive hyperspectral imaging with side information.
\newblock \emph{IEEE Journal of selected topics in Signal Processing}, 9\penalty0 (6):\penalty0 964--976, 2015.

\bibitem[Zhang et~al.(2024)Zhang, Zeng, Chen, Yu, and Zhao]{SSR}
Jiancheng Zhang, Haijin Zeng, Yongyong Chen, Dengxiu Yu, and Yin-Ping Zhao.
\newblock Improving spectral snapshot reconstruction with spectral-spatial rectification.
\newblock In \emph{Proceedings of the IEEE/CVF Conference on Computer Vision and Pattern Recognition (CVPR)}, pages 25817--25826, 2024.

\bibitem[Zhu et~al.(2024{\natexlab{a}})Zhu, Liao, Zhang, Wang, Liu, and Wang]{vision_mamba}
Lianghui Zhu, Bencheng Liao, Qian Zhang, Xinlong Wang, Wenyu Liu, and Xinggang Wang.
\newblock Vision mamba: Efficient visual representation learning with bidirectional state space model.
\newblock \emph{arXiv preprint arXiv:2401.09417}, 2024{\natexlab{a}}.

\bibitem[Zhu et~al.(2024{\natexlab{b}})Zhu, Liao, Zhang, Wang, Liu, and Wang]{vmamba2}
Lianghui Zhu, Bencheng Liao, Qian Zhang, Xinlong Wang, Wenyu Liu, and Xinggang Wang.
\newblock Vision mamba: Efficient visual representation learning with bidirectional state space model.
\newblock \emph{arXiv preprint arXiv:2401.09417}, 2024{\natexlab{b}}.

\end{thebibliography}
}



\end{document}